\RequirePackage{fix-cm}
\documentclass[smallcondensed]{svjour3}     
\smartqed  
\usepackage{graphicx}
\graphicspath{ {./pic_line/} }
 \usepackage{mathptmx}      
\usepackage[numbers]{natbib}
\usepackage{amssymb}
\usepackage{amsmath}
\usepackage[linesnumbered,ruled,vlined]{algorithm2e}
\usepackage{caption}
\usepackage{subcaption}
\usepackage{bm}
\usepackage{multirow}
\usepackage{xcolor}
\newcommand{\revOne}[1]{{\leavevmode\color{black}#1}}

 \journalname{Applied Intelligence}
\begin{document}

\title{GOALS: Gradient-Only Approximations for Line Searches Towards Robust and Consistent Training of Deep Neural Networks
}


\author{Younghwan Chae         \and
        Daniel N. Wilke \and 
        Dominic Kafka  
}


\institute{Y. Chae \at
              University of Pretoria, Lynnwood Rd, Hatfield, Pretoria, 0002 \\
              \email{u11085160@tuks.co.za}           
           \and
           D.N. Wilke \at
              University of Pretoria, Lynnwood Rd, Hatfield, Pretoria, 0002 \\
\email{nico.wilke@up.ac.za}           
           \and
           D. Kafka \at
           University of Pretoria, Lynnwood Rd, Hatfield, Pretoria, 0002 \\
           \email{u11207312@tuks.co.za}
}

\date{Received: date / Accepted: date}

\maketitle

\begin{abstract}
Mini-batch sub-sampling (MBSS) is favored in deep neural network training to reduce the computational cost. Still, it introduces an inherent sampling error, making the selection of appropriate learning rates challenging. The sampling errors can manifest either as a bias or variances in a line search. Dynamic MBSS re-samples a mini-batch at every function evaluation. Hence, dynamic MBSS results in point-wise discontinuous loss functions with smaller bias but larger variance than static sampled loss functions. However, dynamic MBSS has the advantage of having larger data throughput during training but requires the complexity regarding discontinuities to be resolved. This study extends the gradient-only surrogate (GOS), a line search method using quadratic approximation models built with only directional derivative information, for dynamic MBSS loss functions. We propose a gradient-only approximation line search (GOALS) with strong convergence characteristics with defined optimality criterion. We investigate GOALS's performance by applying it on various optimizers that include \textsc{SGD},  \textsc{RMSprop} and \textsc{Adam} on ResNet-18 and EfficientNetB0. We also compare GOALS's against the other existing learning rate methods. We quantify both the best performing and most robust algorithms. For the latter, we introduce a relative robust criterion that allows us to quantify the difference between an algorithm and the best performing algorithm for a given problem. The results show that training a model with the recommended learning rate for a class of search directions helps to reduce the model errors in multimodal cases.
\keywords{Line search \and Learning rate \and Approximation model \and Stochastic gradient \and SNN-GPP}
\end{abstract}

\section{Introduction}
\label{intro}
In neural network training, choosing appropriate learning rates or learning rate schedules is non-trivial\cite{bengio2012practical, Goodfellow2016, strubell2019energy}. As the neural network architectures become larger and more complex, the cost of training increases significantly \cite{NEURIPS2020_1457c0d6}. The monetary, energy and CO2 emission consequence of selecting a training strategy with significant performance variance, i.e. having near-optimal or poor training performance when combined with various optimizers, neural network architectures, and datasets, is noteworthy and important. Therefore, it becomes more and more important to formally quantify the robustness and consistency of a training approach instead of only quantifying its best performance. This also highlights the inherent Pareto optimal nature of selecting a training approach optimal for a specific application (specialist) or being adequate over a larger domain of applications (generalist). In other words, the learning rate strategy selected needs to be robust enough that when it performs sub-optimally, the difference between its performance and the best performing approaches is limited. Incorporating this as a formal selection criterion improves the certainty with which an analyst can interpret the performance of an algorithm on a given problem without having to conduct additional exhaustive studies.

A natural consideration to resolve learning rates may be to consider line searches. 
Line searches are well-established in mathematical programming to efficiently resolve learning rates and identify descent directions. However, they require the underlying loss function to be convex or unimodal over an identified interval. This would be the case if full-batch training of machine learning and deep learning neural networks would be attenable. However, when conducting full-batch training, computational and memory requirements are untenable for practical training. This makes full-batch training ill-suited for DNNs on modern memory limited graphics processing unit (GPU) compute devices. As a result, the standard training procedure for machine learning and deep learning relies on mini-batch sub-sampling.


Mini-batch sub-sampling (MBSS) reduces the computational cost by using only a sub-sample of the training data at a time. 
This also provides a generalization effect \citep{masters2018revisiting} by turning a smooth continuous optimization problem into a stochastic optimization problem \citep{robbins1951stochastic}. 
The stochastic or discontinuous nature of the loss function is due to the selected mini-batches' inherent sampling errors.

\begin{figure}
	\centering
	\begin{subfigure}[b]{0.45\textwidth}
		\includegraphics[width=\textwidth]{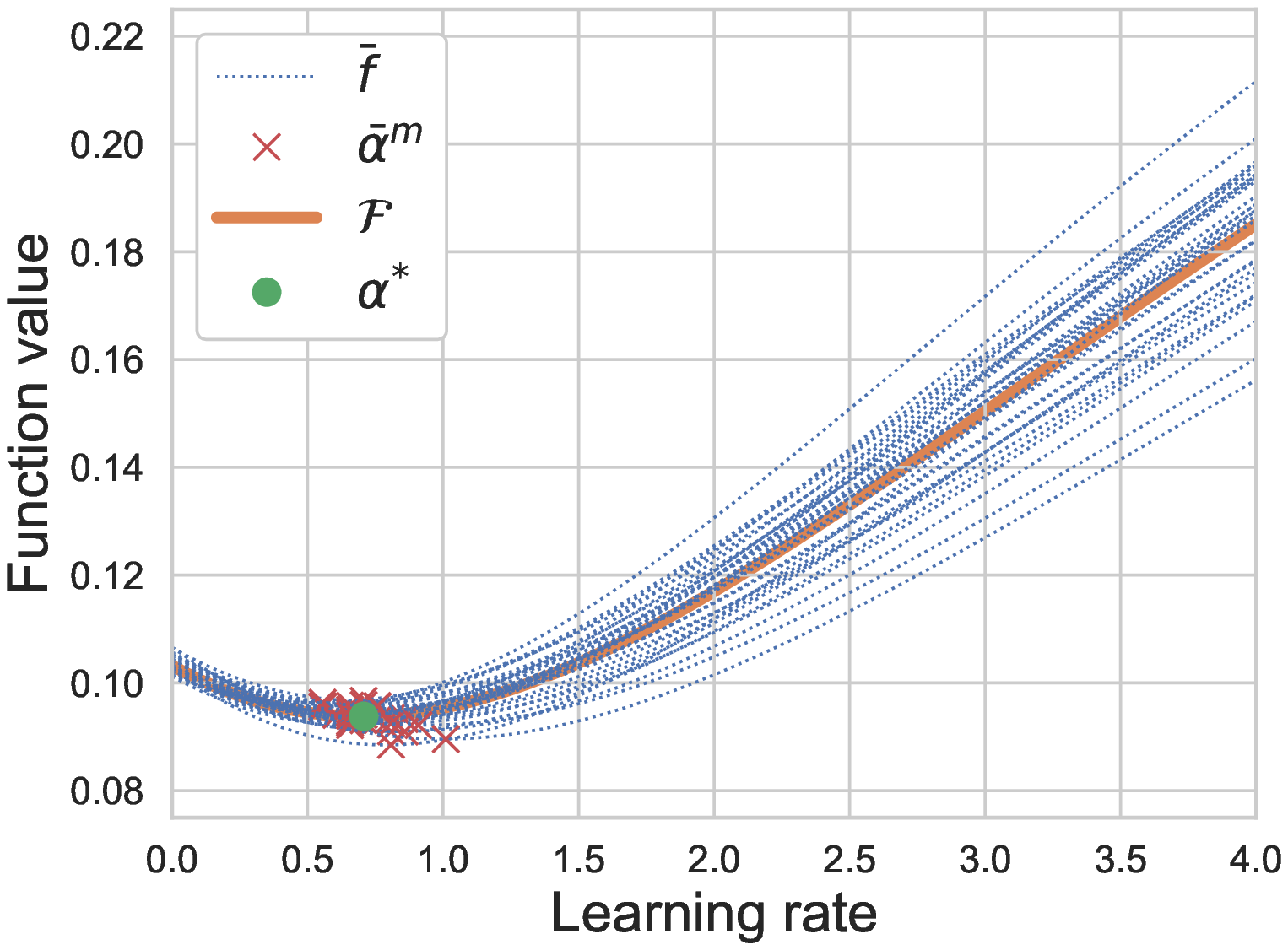}
		\caption{Minimization with static MBSS}
	\end{subfigure}
	\centering
	\begin{subfigure}[b]{0.45\textwidth}
		\includegraphics[width=\textwidth]{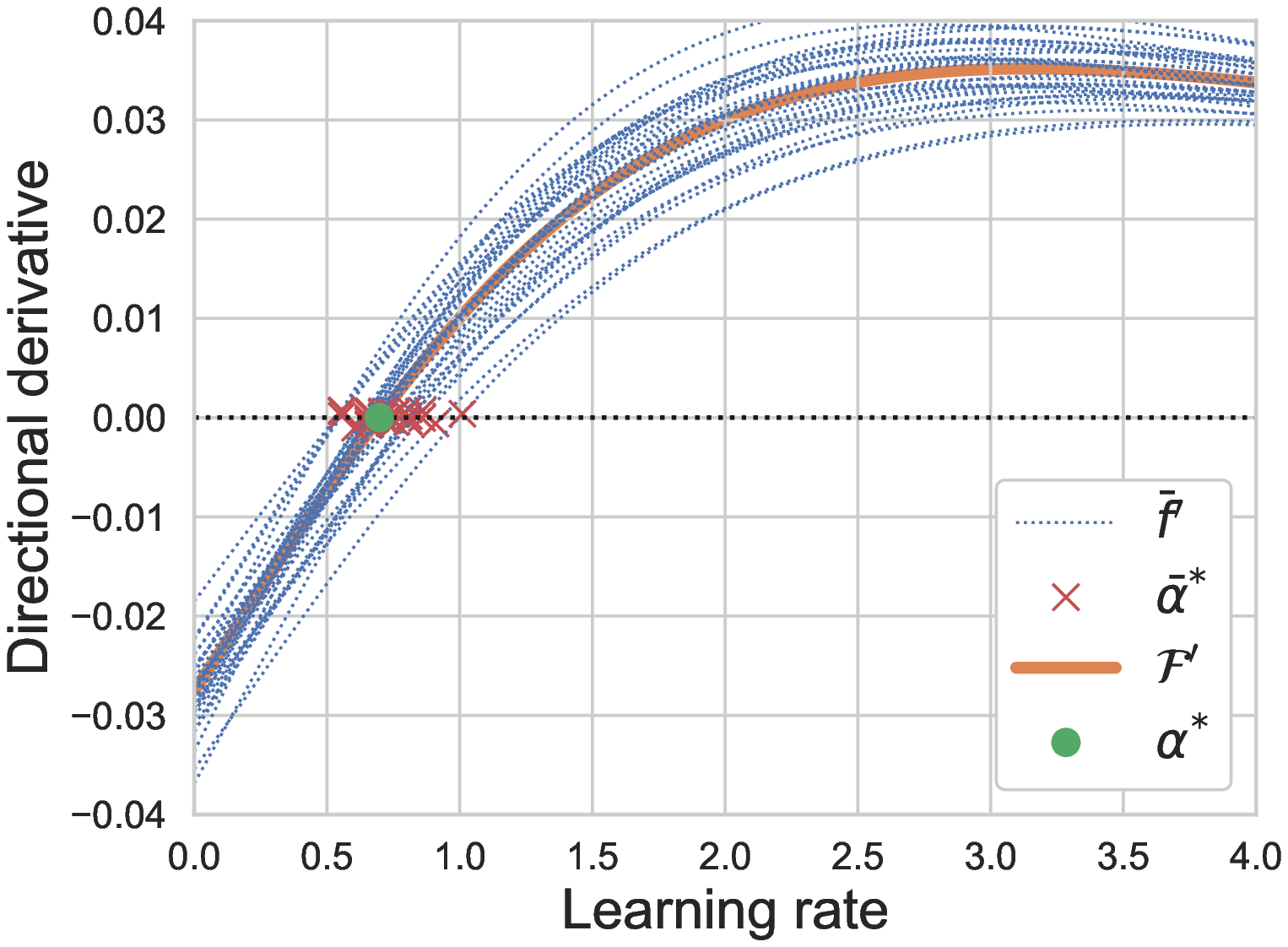}
		\caption{Locating SNN-GPPs with static MBSS}
	\end{subfigure}
	\centering
	\begin{subfigure}[b]{0.45\textwidth}
		\includegraphics[width=\textwidth]{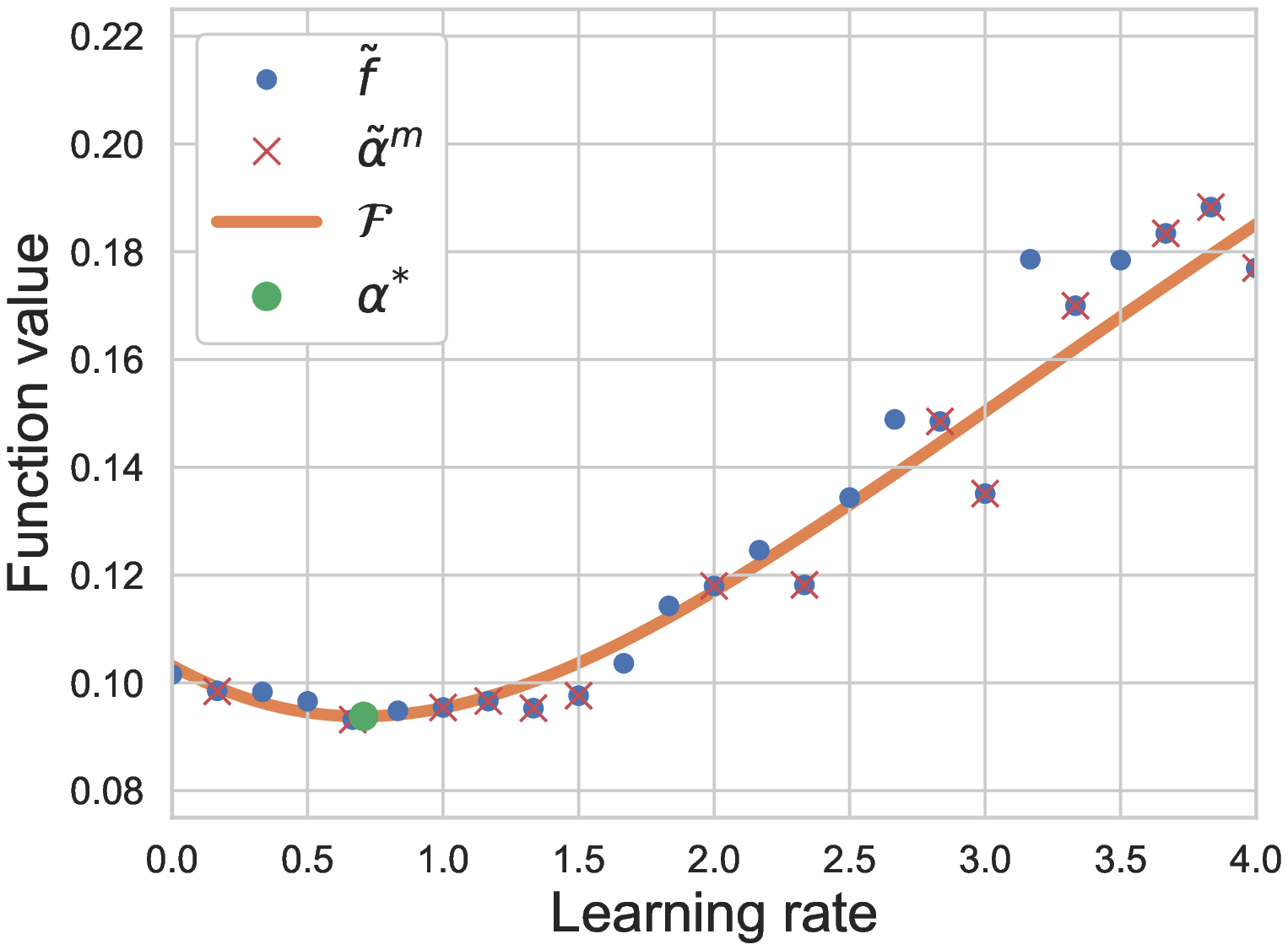}
		\caption{Minimization with dynamic MBSS}
	\end{subfigure}
	\centering
	\begin{subfigure}[b]{0.45\textwidth}
		\includegraphics[width=\textwidth]{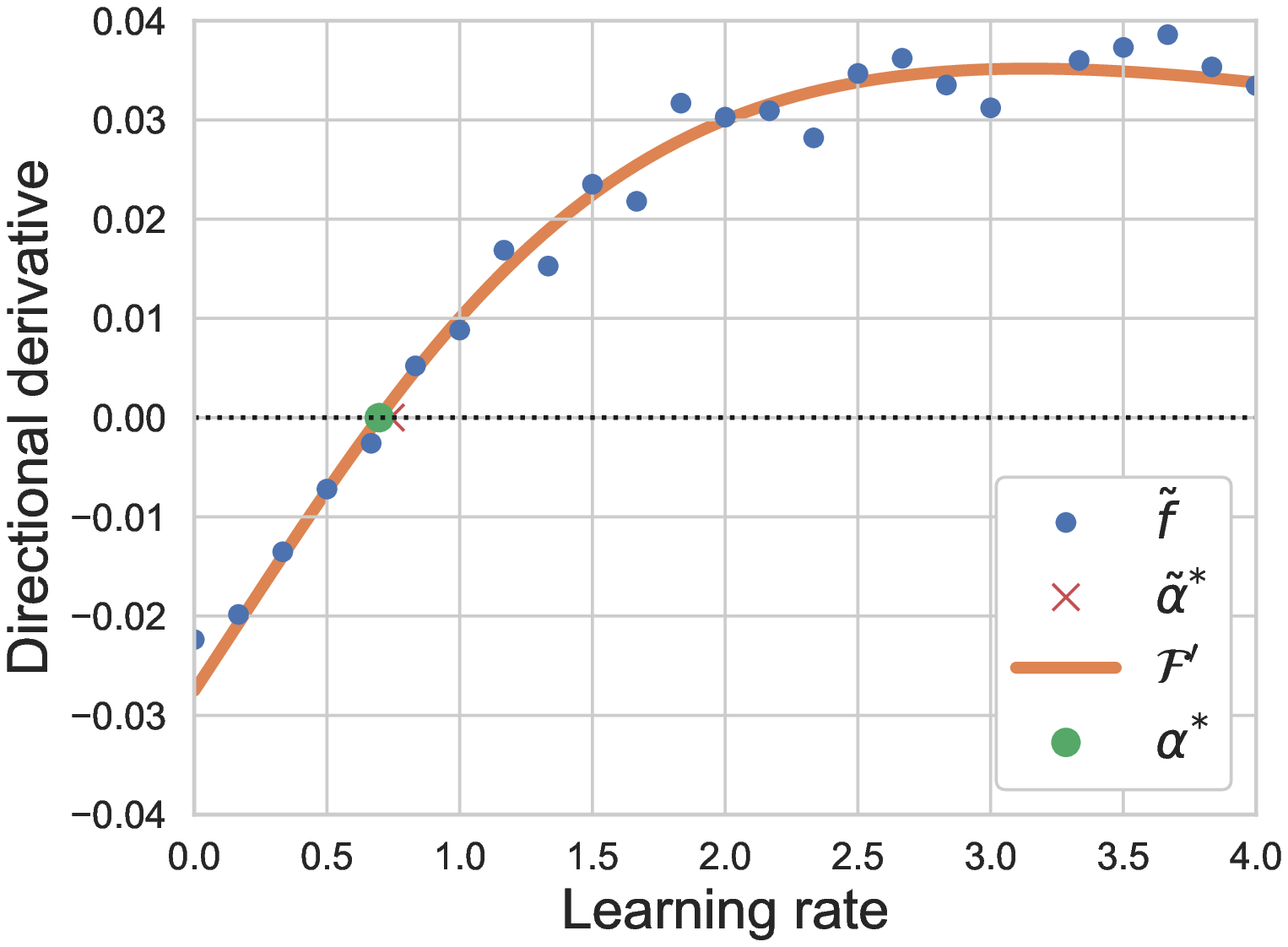}
		\caption{Locating SNN-GPPs with dynamic MBSS}
	\end{subfigure}
	\caption{
		Illustration of finding local minima in (a) static and (c) dynamic MBSS loss functions, as well as locating SNN-GPPs using (b) static and (d) dynamic MBSS directional derivatives.
	}
	\label{MBSS}
\end{figure}

For line searches, the sampling errors manifest mainly in the form of bias or variance along a descent direction, depending on whether mini-batches are sub-sampled statically or dynamically \citep{chae2019empirical, Kafka2019jogo}. 
Static MBSS sub-samples a new mini-batch for every descent direction, while dynamic MBSS sub-samples a new mini-batch for every function evaluation. 
Hence, the loss function for static MBSS is continuous along a descent direction. 
The consequence is that the expected value of a static MBSS loss, has a small variance but a large bias compared to the expected or full-batch response. 
Conversely, the loss function for dynamic MBSS is point-wise discontinuous \citep{kafka2019gradient, chae2019empirical}. 
The expected response of the dynamic MBSS loss has a small bias but a large variance compared to the expected or full-batch response \citep{kafka2019gradient, chae2019empirical}. 
Consider Figure \ref{MBSS}, which contrasts a full batched sampled loss function ($\mathcal{F}$ - orange) against a static ($\bar{f}$) and dynamic ($\tilde{f}$) sampled loss functions for fully-connected feedforward neural network \citep{mahsereci2017probabilistic} initialized with Xavier initialization \citep{glorot2010understanding}. 
Figure  \ref{MBSS} (a) depicts 20 potential static MBSS loss functions. 
Each loss  has zero variance but a large bias in this case. 
Figure  \ref{MBSS} (c) depicts a dynamic MBSS loss function. 
It is clear that there is a large variance in the loss response, but the expected response has a lower bias since a mini-batch does not influence it in particular.

Line searches have been implemented for both static MBSS and dynamic MBSS loss functions. 
The main drawback of static MBSS is that it results in large biases in loss approximations and has been improved by applying sample variance reduction techniques \citep{friedlander2012hybrid, bollapragada2018adaptive}. 
Meanwhile, attempts to resolve learning rates in the point-wise discontinuous loss approximations of dynamic MBSS include the probabilistic line search \citep{mahsereci2017probabilistic}  and Gradient-Only Line Search that is Inexact (GOLS-I)  \citep{kafka2019gradient}. 
The probabilistic line search resolve learning rates by minimizing an approximation constructed using both function value and directional derivative information, while GOLS-I uses only directional derivative sign change information.

GOLS-I locates optima by searching for stochastic non-negative gradient projection points (SNN-GPP). These manifest as positive directional derivatives with a non-zero probability around a ball encapsulating SNN-GPPs. Any point around the SNN-GPP ball is taken and the directional derivative from the SNN-GPP to the point computed. An SNN-GPP along a descent direction merely manifests as a sign change from negative to positive. Importantly, a sign change from negative to positive is necessary and sufficient to identify an SNN-GPP, or alternatively stated, a minimizer as inferred solely from derivative information for this univariate case. This is empirically demonstrated in Figure \ref{MBSS}. 

Learning rates resolved by locating minimizers or SNN-GPPs are equivalent for static MBSS loss functions as depicted by the minimizer solution, $ \bar{\alpha}^{m} $, and SNN-GPP solution, $\bar{\alpha}^*$, in Figures \ref{MBSS} (a) and (b), respectively. However, for dynamic sampled loss functions, minimizers are identified over the entire domain, as shown in Figure \ref{MBSS} (c). In turn, SNN-GPPs are concentrated around the full batch solution, as shown in Figure \ref{MBSS} (d). SNN-GPPs indicate a lower bias to the expected full-batch minimizer than minimizers of SNN-GPPs resolved from the static MBSS loss.

A recent empirical study investigated how function value and directional derivative information help locate SNN-GPPs by comparing the resulting learning rates from various quadratic approximation models built from enforcing information in multiple ways \citep{chae2019empirical}.  
\cite{chae2019empirical} demonstrated that using only function value information resulted in learning rates with the larger variance due to the larger variance when predicting function values when conducting dynamic MBSS. 
Using only derivative information resulted in learning rates with smaller variance as the derivative information predicts more consistently when considering dynamic MBSS. 
Hence, gradient-only quadratic approximations result in stable and consistent learning rate predictions. 
\cite{chae2019empirical} constructed derivative only approximations using only two directional derivative evaluations referred to as gradient-only surrogates (GOS). 
One evaluated at the origin, and the other at an ``initial guess" learning rate along the descent direction. 
The directional derivative at the origin is strictly less than zero for descent directions. 
If the directional derivative at the initial guess is also less than zero, the initial guess learning rate is immediately accepted. 
In turn, if the directional derivative at the initial guess is positive, linear interpolation between the two points is performed to approximate the location of a directional derivative sign change. 
The proposed approach served as an initial investigation and proof of a ``vanilla” line-search concept for only stochastic gradient descent (SGD)
The ``vanilla” directional-derivative-only approximation line search, proposed and investigated by \cite{chae2019empirical}, has no strong convergence characteristics, lacks a robust bracketing strategy and has not been demonstrated for descent directions strategies other than SGD. 
The contributions of this study include:
\begin{enumerate}
	\item It extends the shortcomings of the vanilla directional-derivative-only approximation line search by proposing a line search with strong convergence characteristics. This is achieved by introducing a robust bracketing strategy to improve linear interpolation accuracy, referred to as the gradient-only approximation line search (GOALS). The bracketing strategy is based on a modified strong Wolfe condition \citep{wolfe1969convergence, wolfe1971convergence} to isolate SNN-GPP. We essentially propose a conservative algorithm with strong convergence characteristics, which may result sacrifice the performance for convergence.
	\item GOALS is demonstrated as a suitable line search strategy for descent direction approaches other than SGD, including  \textsc{RMSprop} and \textsc{Adam} on deep neural network architectures with CIFAR-10 \citep{Krizhevsky2009}.
	\item GOALS is compared to fixed learning rates, cosine annealing, GOLS-I, and GOS line search strategies on a shallow neural network architecture with MNIST \citep{lecun1998gradient}. GOALS exhibits competitive results compared to other line search methods.
\end{enumerate}

\section{Background}
In general, line searches can be employed to train deep neural networks to identify minimizers, first-order optimality candidate solutions (directional derivatives equal to 0) and SNN-GPPs. 
For convex functions, all three are equivalent. 
Several line searches which implement static MBSS have been presented, which take advantage of continuous loss functions that often assume convexity \citep{ friedlander2012hybrid, byrd2011use, Byrd2012,bollapragada2018adaptive, kungurtsev2018algorithms, bergou2018subsampling, mutschler2019parabolic,yedida2021lipschitzlr}. 

However, as illustrated in Figure \ref{MBSS}, for dynamic MBSS loss functions, SNN-GPPs identify sensible solutions when compared to the full batch solution. Minimizers are hampered by local minima resulting in large variance, while first-order optimality candidate solutions may not exist for point-wise discontinuous loss functions. The present section summarizes several state-of-the-art sub-sampling and line search schemes applied to dynamic MBSS loss functions in machine learning literature. Firstly, we formalize dynamic MBSS and SNN-GPPs in Sections~\ref{sec:dmbss} and \ref{sec:sngpp}, respectively.

\subsection{Dynamic mini-batch sub-sampling}
\label{sec:dmbss}

Given weights $\boldsymbol{x}$, the function value computed with dynamic MBSS is expressed as
\begin{equation}\label{key}
	\tilde{L}(\boldsymbol{x}) =\dfrac{1}{|\mathcal{B}_{n,i}|}\sum_{b\in\mathcal{B}_{n,i}}\ell(\boldsymbol{x};\boldsymbol{t}_{b}),
\end{equation}
where $ \ell(\boldsymbol{x};\boldsymbol{t}_{b}) $ is computed using training samples in the sampled mini-batch,  $\boldsymbol{t}_{b} $, with
approximate gradient given by
\begin{equation}\label{key}
	\tilde{\boldsymbol{g}}(\boldsymbol{x})= \dfrac{1}{|\mathcal{B}_{n,i}|}\sum_{b\in\mathcal{B}_{n,i}}\bm{\nabla}\bm{\ell}(\boldsymbol{x};\boldsymbol{t}_{b}),
\end{equation}
where $ i $ denotes the $ i $-th function evaluation of the $ n $-th iteration of a given algorithm. The loss function as a function of learning rate, $ \alpha $, along a given descent direction, $ \boldsymbol{d}_{n} $, starting from $ \boldsymbol{x}_{n} $ is given by:
\begin{equation}\label{dynamic1}
	\tilde{f}_{n}(\alpha) =\tilde{L}(\boldsymbol{x}(\alpha)) = \tilde{L}(\boldsymbol{x}_{n} + \alpha\boldsymbol{d}_{n}),
\end{equation}
with the directional derivative, $ \tilde{f}'_{n} $, given by
\begin{equation}\label{key}
	\tilde{f}_{n}'(\alpha) = \boldsymbol{d}_{n}^{\intercal}\tilde{\boldsymbol{g}}(\boldsymbol{x}_{n}+ \alpha\boldsymbol{d}_{n}).
\end{equation}
Dynamic MBSS loss functions are point-wise discontinuous functions with point-wise discontinuous gradient fields.

\subsection{Gradient-only optimality criterion}
\label{sec:sngpp}
Multiple local minima would be found when locating minimisers for discontinuous functions such as a dynamic MBSS loss function.
Instead, we may opt to locate Non-Negative Gradient Projection Points (NN-GPPs) for which its gradient-only optimality criterion was specifically designed for deterministic discontinuous function \citep{wilke2013gradient}, given by
\begin{equation}\label{nngpp}
	\boldsymbol{d}_{n}^{\intercal} \bm{\nabla} \bm{\mathcal{L}}(\boldsymbol{x}_{nngpp} + \alpha_{n}\boldsymbol{d}_{n}) \geq 0, \quad \forall \|\boldsymbol{d}_{n} \in \mathbb{R}^{p}\|_{2} = 1, \quad \forall\alpha \in (0, \alpha_{max}],
\end{equation}
for the 1-D case, along a given search direction, $\boldsymbol{d}_n$. NN-GPP is representative of a local optimum because no descent directions are allowed away from it. This is only possible at a critical point or a local minimum in a smooth and continuous function.

The NN-GPP definition is limited to deterministic discontinuous functions. Therefore, to accommodate stochastic discontinuous functions, the NN-GPP definition was generalized and extended to the Stochastic NN-GPP (SNN-GPP), given by
\begin{equation}\label{snngpp}
	\boldsymbol{d}_{n}^{\intercal}\tilde{\boldsymbol{g}}(\boldsymbol{x}_{snngpp} + \alpha_{n}\boldsymbol{d}_{n}) \geq 0, \enspace \forall \|\boldsymbol{d}_{n} \in \mathbb{R}^{p}\|_{2} = 1, \enspace \forall\alpha \in (0, \alpha_{max}], \enspace p(\boldsymbol{x}_{snngpp})>0,
\end{equation}
with probability, $ p(\boldsymbol{x}_{snngpp}) $, greater than 0 \citep{kafka2019gradient}.

The difference between NN-GPP and SNN-GPP is that NN-GPP is a point where the signs of directional derivatives change in the deterministic setting. However, in the stochastic setting, a directional derivative sign change location may vary, depending on the instance of the sampled stochastic loss.
Transferred to dynamic MBSS losses, this means that for each distinct mini-batch, $ \mathcal{B} $, selected, we have a distinct location of a sign change. However, these remain bounded in a ball, $B_\epsilon$, \citep{kafka2019gradient} of a given loss landscape neighborhood. The size of $B_\epsilon$ is, among other factors, dependent on the variance in the stochastic loss function, which in dynamic MBSS losses is dependent on the mini-batch size. Hence, the larger the difference between individual samples, $ \mathcal{B}_{n,i} $, the larger the size of the ball, $B_\epsilon$. Notably, the SNN-GPP definition also generalizes to the NN-GPP, critical point and local minimum, as these are all SNN-GPPs with probability 1.

\subsection{Line searches for dynamic MBSS loss functions}

To the best of the author's knowledge, only three line search techniques have been proposed to resolve learning rates for dynamic MBSS loss functions, namely:
\begin{enumerate}
	\item A probabilistic line search using Bayesian optimization with Gaussian surrogate models, built using both function value and directional derivative information \citep{mahsereci2017probabilistic}.
	\item The Gradient-only line search, Inexact (GOLS-I), locates SNN-GPPs along the search directions, using only directional derivative information \citep{kafka2019gradient}.
	\item Proof of concept quadratic approximations \citep{chae2019empirical}.
\end{enumerate}

All three line search methods showed competitive training performance for dynamic MBSS losses and outperformed various constant step sizes. 

Notably, \cite{chae2019empirical} the quality of function values and directional derivatives in the context of approximation-based line searches is investigated empirically. 
The quality of information used to produce approximations in dynamic MBSS losses was studied by constructing five types of quadratic approximation models using different information sampled at two locations (e.g. only function value, only directional derivative, both function value and directional derivative models) \citep{chae2019empirical}. 
The results showed that using directional derivative information at the origin (starting point) of a line search is critical for constructing quality approximations, decreasing the variances in optimal learning rates. 
The two best performing models were
\begin{enumerate}
	\item Derivative-only quadratic model: A quadratic approximation is built using two directional derivatives, values measured at the origin and another point along descent direction.
	\item Mixed quadratic model: A quadratic approximation is built using the directional derivative measured at the origin and function values at both origin and another point.
\end{enumerate}
Note that both quadratic approximation models proposed by \cite{chae2019empirical} demonstrate ``vanilla" algorithms without guaranteed convergence. Although the mixed-model has been investigated by \cite{mahsereci2017probabilistic} and \cite{mutschler2019parabolic} before, the derivative-only model has not been extended to a fully automated line search technique, which is the aim of this paper. Next, we discuss the heuristics and the corresponding shortcomings of the vanilla derivative-only approximation using the derivative-only model proposed in \cite{chae2019empirical}, extending in this study.

\subsection{Gradient only surrogate (GOS)}\label{model}
Given a multivariate dynamic MBSS loss function, $ \tilde{L}(\boldsymbol{x}_{n})$, along a given descent direction, $ \boldsymbol{d}_{n}$, $ \tilde{f}(\alpha) $, we want to resolve the learning rate, $ \alpha $. The quadratic approximation model, $ \hat{f}(\alpha) $, of $ \tilde{f}(\alpha) $ is given by
\begin{equation}\label{quad_eq}
	\hat{f}(\alpha) := k_{1}\alpha^{2} + k_{2}\alpha + k_{3} \approx \tilde{f}(\alpha),
\end{equation}
where $ k_{1}, k_{2} $ and $ k_{3} $ are the constants to be computed. Similarly, the first-order derivative of the quadratic approximation, $ \hat{f}'(\alpha) $, is a linear approximation given by
\begin{equation}\label{lin_eq}
	\hat{f}'(\alpha) := 2k_{1}\alpha + k_{2} \approx \tilde{f}'(\alpha).
\end{equation}
Note that $ \hat{f}'(\alpha) $ is the derivative-only approximation proposed by \cite{chae2019empirical}, and is implemented throughout this paper. 
The approximation uses only directional derivative information. 
The constants at $ n $-th iteration, $ k_{1,n} $ and $ k_{2,n} $ can be solved using a linear system of equations, constructed from two instances of Equation~\eqref{lin_eq}, given by
\begin{equation}\label{linsys}
	\begin{bmatrix}
		2\alpha_{0,n} & 1 \\
		2\alpha_{1,n} & 1
	\end{bmatrix}
	\begin{bmatrix}
		k_{1,n} \\ k_{2,n}
	\end{bmatrix}
	=
	\begin{bmatrix}
		\tilde{f}'_{0,n}\\
		\tilde{f}'_{1,n}
	\end{bmatrix},
\end{equation}
where $ \tilde{f}'_{0,n} $ and $ \tilde{f}'_{1,n} $ denote the dynamic MBSS directional derivatives measured at $ \alpha_{0,n} $ and $ \alpha_{1,n} $ respectively, where $ \alpha_{0,n} =0 $ is the current starting point and $ \alpha_{1,n} $ is the initial guess ($ \alpha_{1,n} > 0 $). The approximate minimum, $ \tilde{\alpha}^{*}_{n} $, is computed as $ \alpha^{*}_{n} = -k_{2,n}/(2k_{1,n})$, where $ \hat{f}'(\tilde{\alpha}^{*}_{n}) = 0 $. For the implementation, we compute $ \tilde{\alpha}^{*}_{n} $ in the closed-form as follows:
\begin{equation}\label{lin_interp}
	\alpha^{*}_{n} = \alpha_{0,n} - \tilde{f}'_{0,n} \frac{\Delta \alpha_{n}}{\Delta \tilde{f}'_{n}} = \alpha_{0,n} - \tilde{f}'_{0,n} \frac{\alpha_{1,n}-\alpha_{0,n}}{\tilde{f}'_{1,n}-\tilde{f}'_{0,n}}.
\end{equation}

\begin{figure}
	\begin{subfigure}[b]{0.30\textwidth}
		\centering
		\includegraphics[width=\textwidth]{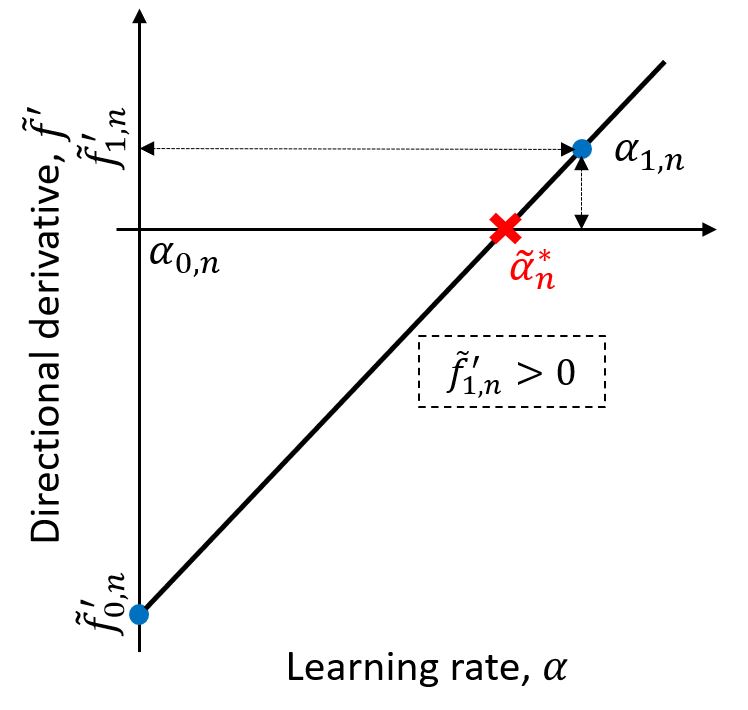}
		\caption{Bounded Interpolation}
	\end{subfigure}
	\begin{subfigure}[b]{0.30\textwidth}
		\centering
		\includegraphics[width=\textwidth]{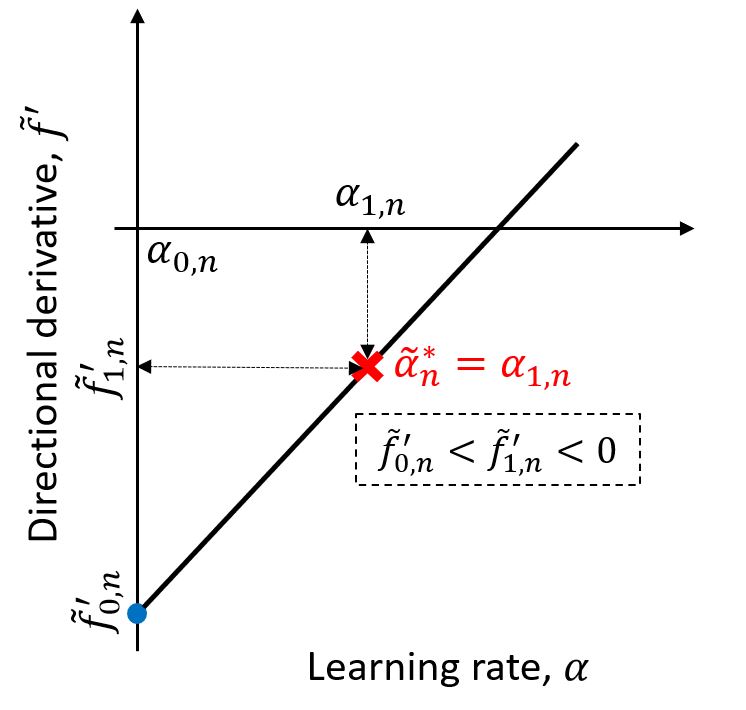}
		\caption{Bounded extrapolation}
	\end{subfigure}
	\begin{subfigure}[b]{0.30\textwidth}
		\centering
		\includegraphics[width=\textwidth]{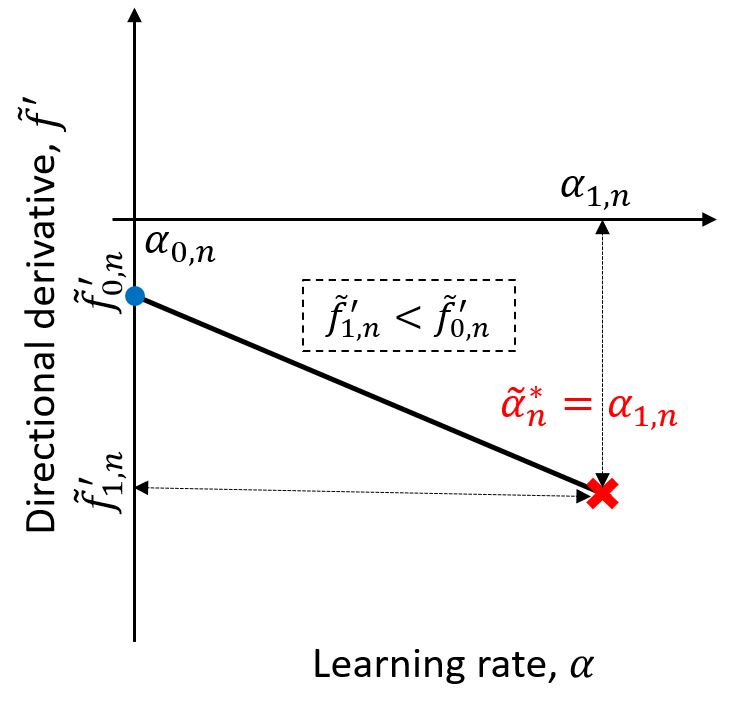}
		\caption{Unbounded extrapolation}
	\end{subfigure}
	\caption{
		Illustration of three possible cases when implementing the vanilla line search algorithm using the derivative-only approximation: (a) bounded interpolation, when $ \tilde{f}'_{1,n} > 0 $, (b) bounded extrapolation when $ \tilde{f}'_{0,n} <\tilde{f}'_{1,n} < 0 $ and (c) unbounded extrapolation, when $ \tilde{f}'_{1,n} < \tilde{f}'_{0,n} $.
	}
	\label{vanilla}
\end{figure}

The heuristics of the vanilla line search algorithm using Equation~\eqref{lin_interp}, proposed by \cite{chae2019empirical}, are recalled here:

\begin{enumerate}
	\item If $ \tilde{f}'_{1,n} > 0 $, as shown in Figure~\ref{vanilla}(a), we may perform bounded linear interpolations, Equation~\eqref{lin_interp}, to resolve learning rate, $ \tilde{\alpha}^{*}_{n} $.
	\item If $ \tilde{f}'_{0,n} <\tilde{f}'_{1,n} < 0 $, as shown in Figure~\ref{vanilla}(b), we can perform bounded extrapolation using  Equation~\eqref{lin_interp}, but the prediction error is expected to be larger than the case of bounded interpolation. Hence, we immediately choose the initial guess as the resulted learning rate, i.e. $ \tilde{\alpha}^{*}_{n} = \alpha_{1,n}$.
	\item If $ \tilde{f}'_{1,n} < \tilde{f}'_{0,n} $, as shown in Figure~\ref{vanilla}(c), it would be unwise to perform unbounded extrapolation for the same reason as in the case of bounded extrapolation. Therefore, we immediately accept the initial guess $ \alpha^{*} $, as the learning rate of the current iteration $ \tilde{\alpha}^{*}_{n} = \alpha_{1,n} $.
\end{enumerate}

The examples, studied by \cite{chae2019empirical}, used stochastic gradient descent (SGD) with the vanilla algorithm, $ \boldsymbol{d}_{n} = -\tilde{\boldsymbol{g}}_{n} $. 
The initial guess, $ \alpha_{1,n} $, for every iteration was chosen to be the inverse of L-2 norm of the search direction vector, $ \alpha_{1,n} = \|\boldsymbol{d}\|_{2}^{-1} $. 
This means that $ \alpha_{1,n} $ is adapted to the magnitude of the search direction vector, $ \boldsymbol{d}_{n} $, to prevent overly aggressive (potentially unstable) training behavior, when $ \| \nabla\tilde{\bm{\mathcal{L}}}_{n}\|_{2} \approx 0$.

\section{\revOne{Robustness measure, $ R $}}
\label{robustness}

As DNN problems have become more complex, we might prefer using a robust optimizer that performs well across various problems to a problem-specific optimizer for training an unseen problem. In other words, we would be interested in the strategy not necessarily outperforms all the other optimizers for only a specific problem but has the slightest differences in performance from the best version of each problem. Hence, we define the relative robustness measure for learning rate strategies as follows:

\begin{definition}[Relative robustness]
	The relative robustness, $ R $, of a strategy, $ y $, is computed by summing the absolute differences, $ \psi_{y,h,o} $, in the strategy's accuracy, $ \eta_{y,h,o} $, and the best accuracy of all strategies, $ \eta_{h,o}^{*} $, for all optimizers, $ O \ni o $, and all problems, $  H \ni h $. Hence, the less the measure, $ R_{y} $, is, the more robust the strategy is. The equation for $ R_{y} $ is given by
	\begin{equation}\label{rr}
		R_{y} = \sum_{h \in H} \sum_{o \in O} \psi_{y,h,o} \text{, where } \psi_{y,h,o} = |\eta_{h,o}^{*}-\eta_{y,h,o}|.
	\end{equation}
\end{definition}

One may also compute a robustness measure, $ R_{y,h} $, for a strategy, $ y $ and a specific problem, $ h $ while considering all optimizers, $ O $. We will compare the training and test performance of our proposed algorithm, GOALS, against the other learning rate strategies based on the relative robustness measure throughout the paper. 

\section{Gradient-only approximation line search (GOALS)}\label{gos}
This section proposes our line search strategy, gradient-only approximation line search (GOALS), using the quadratic derivative-only approximation, capable of locating the SNNGPPs in the stochastic loss functions produced by dynamic MBSS. Unlike the vanilla algorithm, GOALS requires consecutive function evaluations to converge to an interval of sign changes for a given descent direction, $ \boldsymbol{d}_{n} $, update.

GOALS is comprised of the two main stages: 1) An immediate accept condition (IAC), and 2) a bracketing strategy. The IAC means that we accept the initial learning rate guess of the $ n $-th iteration, $ \alpha_{1,n} $, as the approximate solution, $ \tilde{\alpha}^{*}_{n} $, when $ \alpha_{1,n} $ falls within the accepted range set by the Wolfe condition \citep{wolfe1969convergence, wolfe1971convergence}. If the IAC is not satisfied, the proposed bracketing strategy is used to locate SNN-GPPs.

\subsection{Immediate accept condition (IAC)}
The IAC aims to save computational cost. When the immediate accept condition (IAC) is satisfied, we immediately accept the initial guess, $ \alpha_{1,n} $, and continue to the next search direction. The IAC is based on the Wolfe condition, consisting of two conditions: 1) Armijo condition and 2) Wolfe curvature condition. The Armijo condition ensures that the function value at the accepted learning rate decreases monotonically as outlined,
\begin{equation}\label{armijo}
	\tilde{f}_{1,n}\leq \tilde{f}_{0,n} + \omega\alpha_{1,n} \tilde{f}'_{0,n}.
\end{equation}
Here, $ \omega $ is a prescribed constant, often very small (e.g. $ \omega=10^{-4} $). Note that the Armijo condition limits the maximum learning rate by disallowing any increase in function value.

The Wolfe curvature condition ensures that the directional derivative at the initial guess, $ f'_{1,n} $, is less than the directional derivative at the starting point, $ f'_{0,n} $. The condition is given by:
\begin{equation}\label{wolfe}
	-\tilde{f}'_{1,n} \leq  -c\tilde{f}'_{0,n},
\end{equation}
where $ c \in (0,1) $ a prescribed curvature constant. The Wolfe curvature condition limits the minimum learning rate based solely on directional derivative information. In contrast, the Armijo condition, Equation~\eqref{armijo}, requires function value information. Hence, it is not suitable for our derivative-only purpose. However, the Armijo condition is essential for limiting the maximum learning rates by not allowing growth in the function value. Therefore, we will implement the strong Wolfe condition as an alternative to Equations~\eqref{armijo} and \eqref{wolfe}:
\begin{equation}\label{strong}
	|\tilde{f}'_{1,n}| \leq  c|\tilde{f}'_{0,n}|.
\end{equation}
As a consequence of applying the strong Wolfe condition, we gain control over preventing overshooting, Equation~\eqref{armijo}, and undershooting, Equation~\eqref{wolfe}, by changing the $ c $ constant.

Although we implement Equation~\eqref{wolfe} as the IAC throughout the paper, note that one could also independently control undershoot and overshoot limits by splitting $ c $ into two positive constants, $ c_{1} $ and $ c_{2} $, respectively, as follows:
\begin{equation}\label{iac_eq}
	c_{1}\tilde{f}'_{0,n}\leq \tilde{f}'_{1,n} \leq  -c_{2}\tilde{f}'_{0,n},  \quad c_{1}, c_{2} \in [0,1).
\end{equation}

\begin{figure}
	\begin{subfigure}[b]{0.45\textwidth}
		\centering
		\includegraphics[width=\textwidth]{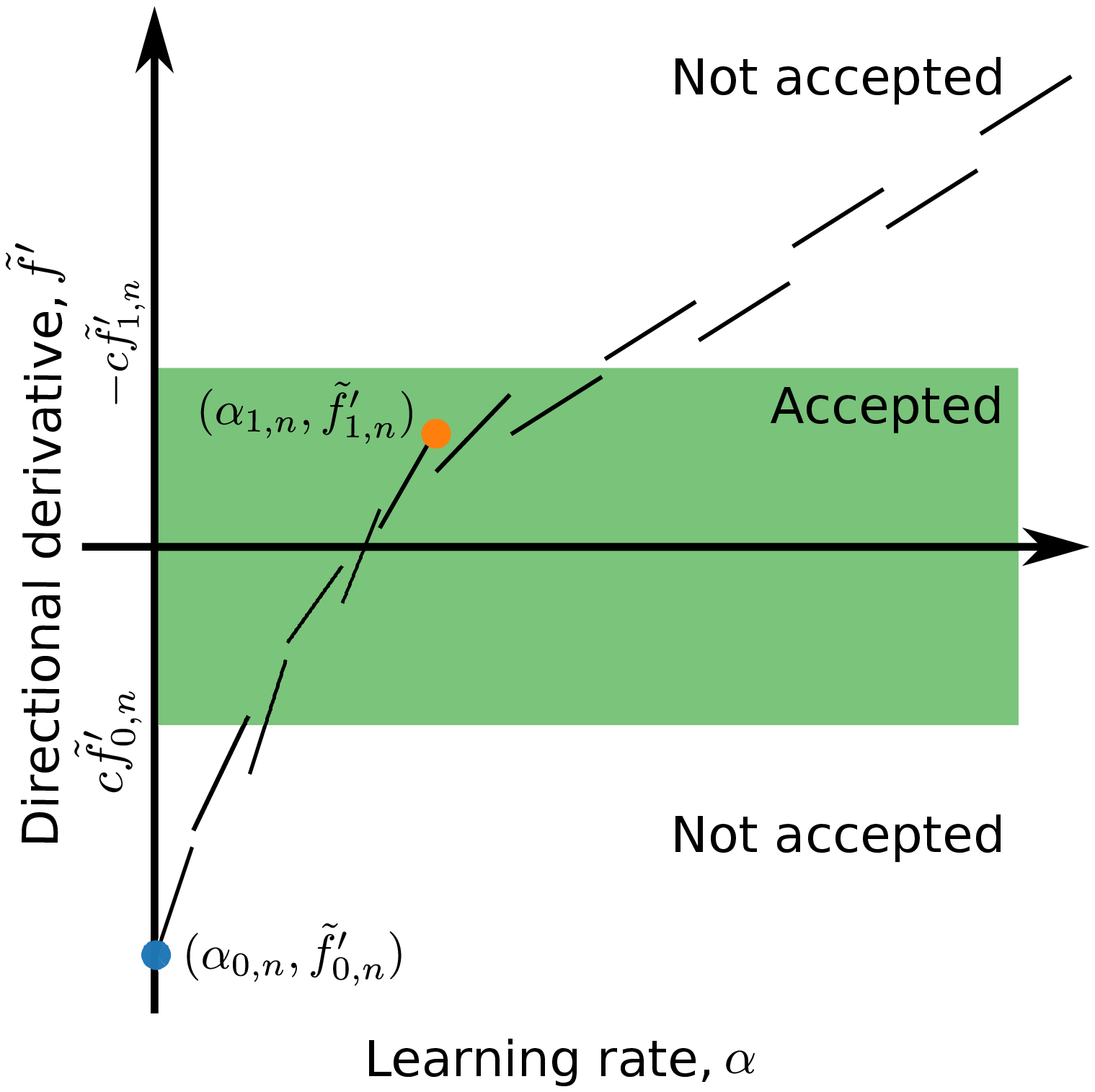}
		\caption{Accepted}
	\end{subfigure}
	\begin{subfigure}[b]{0.45\textwidth}
		\centering
		\centering
		\includegraphics[width=\textwidth]{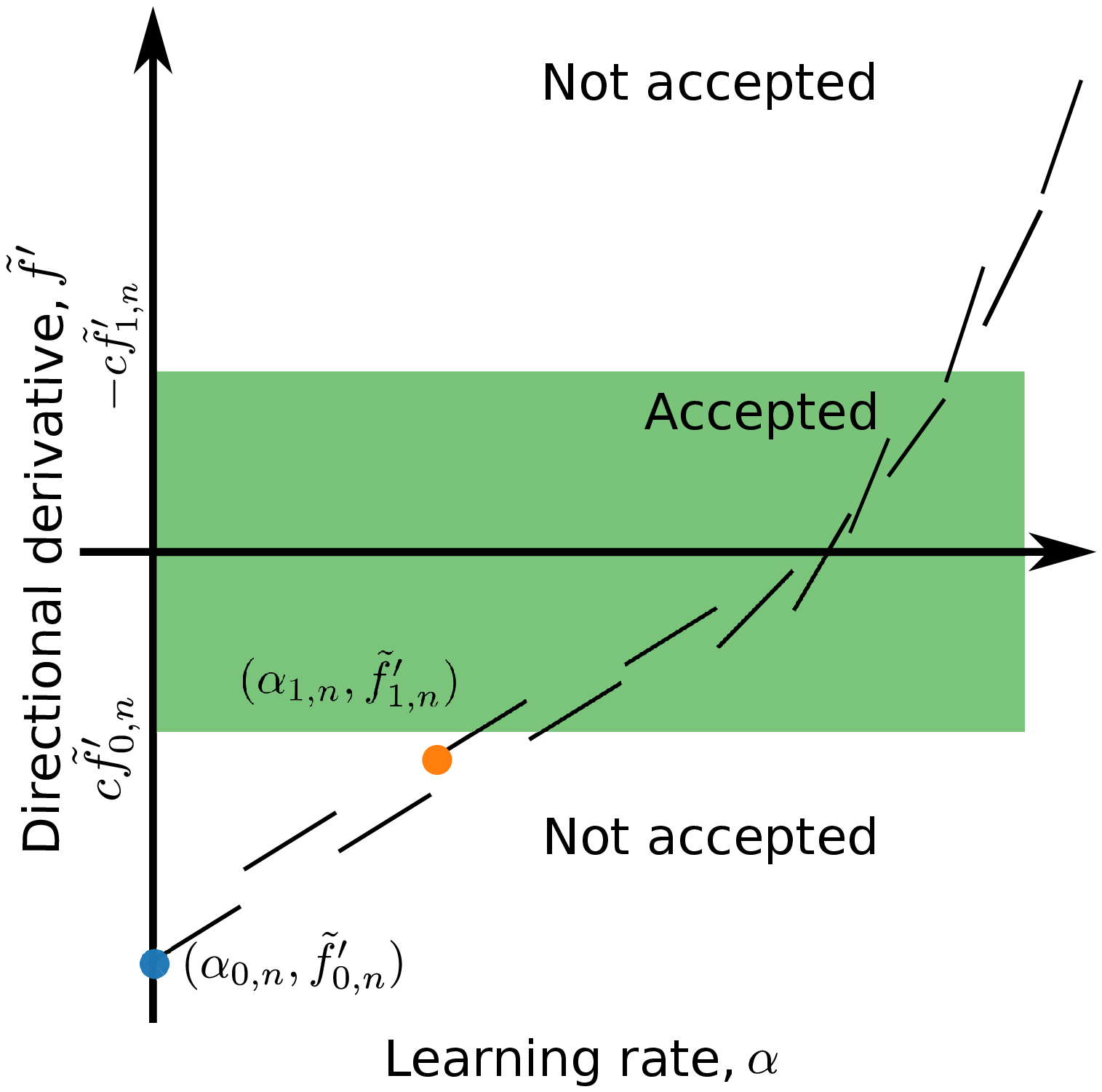}
		\caption{Not accepted}
	\end{subfigure}
	\caption{Illustration of immediate accept condition: (a) when the IAC (\ref{iac_eq}) satisfies, the initial guess, $ \alpha_{1,n} $, is accepted and (b) when the IAC (\ref{iac_eq}) does not satisfy, the initial guess, $\alpha_{1,n} $, is not accepted.}
	\label{iac}
\end{figure}

If the IAC in Equation~\eqref{strong} is satisfied, there is just one directional derivative computation required to determine the learning rate along a descent direction, $ \boldsymbol{d}_{n} $. Figure~\ref{iac} illustrates cases of when the initial guess, $ \tilde{f}'_{1,n} $, is either accepted  (left) for satisfying the IAC condition, or not accepted (right) for not satisfying the condition.

However, if the initial guess is not accepted, we employ the bracketing strategy introduced in the next section to search for SNN-GPPs. Note that the larger the $ c $ value becomes, the larger the range of the IAC becomes. Algorithm \ref{main_alg} lists the pseudocode for the main GOALS algorithm with the IAC.

\begin{algorithm}[H]
	\label{main_alg}
	\DontPrintSemicolon 
	\KwIn{$ \tilde{\boldsymbol{g}}_{0,n} = \tilde{\boldsymbol{g}}^{*}_{n-1} $, $ \tilde{\alpha}^{*}_{n-1} $, $ \boldsymbol{d}_{n} $, $ c $, $ \alpha_{min} $, $ \alpha_{max} $, $ \gamma $ and $ \varepsilon $}
	\KwOut{$ \tilde{\alpha}^{*}_{n} $, $ \tilde{\boldsymbol{g}}^{*}_{n} $}
	Compute directional derivative at $ \alpha_{0,n} \rightarrow \tilde{f}'_{0,n}$ \;
	\tcc{Magnitude check for avoiding numerical issues}
	\If{$ |\tilde{f}'_{0,n}|  < \varepsilon$ \label{alg_error}}
	{
		Recompute $ \tilde{\boldsymbol{g}}^{*}_{n}$ for the next iteration.\;
		\Return $ \tilde{\alpha}^{*}_{n} := 0 $, $\tilde{\boldsymbol{g}}^{*}_{n}$
	}
	\tcc{Setting the initial guess, $ \alpha_{1,n} $}
	$\alpha_{1,n} \rightarrow \gamma$\; \label{initial}
	Compute gradient and directional derivative at $ \alpha_{1,n} \rightarrow  \tilde{\boldsymbol{g}}_{1,n}, \tilde{f}'_{1,n}$\;
	\tcc{Check whether the IAC satisfies}
	\uIf{$ |\tilde{f}'_{1,n}| \leq  c|\tilde{f}'_{0,n}|$ \label{iac_code}}{
		$ \tilde{\alpha}^{*}_{n} := \alpha_{1,n} $ and $ \tilde{\boldsymbol{g}}^{*}_{n} := \tilde{\boldsymbol{g}}_{1,n}$\;}
	\Else{
		$ \tilde{\alpha}^{*}_{n}$,$ \tilde{\boldsymbol{g}}^{*}_{n} :=  Bracketing( \alpha_{0,n} ,  \alpha_{1,n} ,  \tilde{f}'_{0,n} ,  \tilde{f}'_{1,n} ) $\;
	}
	\Return{$ \tilde{\alpha}^{*}_{n}$,$ \tilde{\boldsymbol{g}}^{*}_{n}$}
	\caption{GOALS}
\end{algorithm}

For every iteration, $ n $, GOALS requires the gradient vector at the starting point, $ \tilde{\boldsymbol{g}}_{0,n} $. Therefore, for $ n > 0 $, the resulting gradient from the previous iteration, $ \tilde{\boldsymbol{g}}^{*}_{n-1} $, can be used for $ \tilde{\boldsymbol{g}}_{0,n} $ in the next iteration. In line \ref{alg_error}, the tolerance value, $ \varepsilon $, ensures that $ \tilde{f}'_{0,n}  $ is a numerically positive value. Otherwise, we recompute the gradient at the same point with the resulted learning rate, $ \tilde{\alpha}^{*}_{n}= 0 $, using a new mini-batch, $ \mathcal{B} $, and continue to the next iteration.

In line \ref{initial}, the initial guess, $ \alpha_{1,n} $, is set to be the default learning rate, $\gamma$, often the recommended learning rate for the chosen optimizer. Next, we compute the gradient vector, $ \tilde{\boldsymbol{g}}_{1,n} $, and directional derivative, $ \tilde{f}'_{1,n} $, at the initial guess, $ \alpha_{1,n} $. In line \ref{iac_code}, if the IAC satisfies, we choose the current learning rate, $ \tilde{\alpha}^{*}_{n} = \alpha_{1,n} $, and the resulting gradient, $ \tilde{\boldsymbol{g}}^{*}_{n} = \tilde{\boldsymbol{g}}_{1,n}$. If the IAC is not satisfied, we implement the bracketing strategy to compute $ \tilde{\alpha}^{*}_{n} $ and $ \tilde{\boldsymbol{g}}^{*}_{n} $. The bracketing strategy aims to minimize the model error using linear interpolation, introduced in the following section.

\subsection{Bracketing strategy}
The bracketing strategy aims to isolate an SNN-GPP inside an interval, $ \mathcal{I} \in [\alpha_{L}, \alpha_{U}] $ with lower bound, $ \alpha_{L}$ and upper bound, $ \alpha_{U} $, by updating $ \mathcal{I} $ repeatedly, until the strong Wolfe condition, Equation~\eqref{iac_eq} of $ \tilde{\alpha}^{*}_{n} $ is satisfied. Once the directional derivative signs at $ \alpha_{L} $ and $ \alpha_{U} $ are found to brackets an SNN-GPP, we reduce the interval by applying the Regula-Falsi method \citep{gupta2019numerical}. This is essentially a consecutive linear interpolation method, until $ \tilde{\alpha}^{*}_{n} $ satisfies Equation~\eqref{iac_eq}. We provide the pseudocode for the bracketing strategy in Algorithm~\ref{main_alg2}.

\begin{algorithm}[H]
	\label{main_alg2}
	\DontPrintSemicolon 
	\KwIn{$ \alpha_{0,n} $, $ \alpha_{1,n} $, $ \tilde{f}'_{0,n} $, $ \tilde{f}'_{1,n} $, $  \boldsymbol{d}_{n} $, $ \alpha_{min} $, $ \alpha_{max} $, $c$ and $ \varepsilon $}
	\KwOut{$ \alpha^{*}_{n} $, $ \tilde{\boldsymbol{g}}^{*}_{n} $}
	Initialize $ \alpha_{L} := \alpha_{0,n} $, $ \tilde{f}'_{L} := \tilde{f}'_{0,n} $, $ \alpha_{U} := \alpha_{1,n} $ and $ \tilde{f}'_{U} := \tilde{f}'_{1,n} $ \label{alg2_init}\;
	\tcc{Shifting the interval to larger learning rates}
	\While{$ (\tilde{f}'_{U} < c\tilde{f}'_{0,n}) \text{ and } (2\alpha_{U} < \alpha_{max}) $ \label{alg2_grow}}
	{
		Update $ \alpha_{L} := \alpha_{U} $ and $ \tilde{f}'_{L} := \tilde{f}'_{U} $\;
		$ \alpha_{U} := 2(\alpha_{U}) $ and recompute $  \tilde{\boldsymbol{g}}_{U}, \tilde{f}'_{U} $\;
	}
	$ \tilde{f}'_{temp} := \tilde{f}'_{U} $ \;
	$ \alpha_{temp} := \alpha_{U} $
	
	\tcc{Shrinking the interval using linear interpolations}
	\While{$ (\tilde{f}'_{temp} > -c\tilde{f}_{0,n}) \text{ and } (\tilde{f}'_{U} \tilde{f}'_{L} < 0) \text{ and } (|\tilde{f}'_{U}-\tilde{f}'_{L}| > \varepsilon)$\label{alg2_interp}}{
		$\alpha_{temp} = \frac{\alpha_{L}\tilde{f}'_{U}-\alpha_{U}\tilde{f}'_{L}}{\tilde{f}'_{U}-\tilde{f}'_{L}} $\;\label{alg2_regula}
		Recompute $ \tilde{f}'_{temp} $ at $\alpha_{temp}$\;
		\If{$\tilde{f}'_{temp}\tilde{f}'_{L}<0$}
		{
			$ \tilde{f}'_{U} := \tilde{f}'_{temp} $\;
			$ \alpha_{U} = \alpha_{temp} $
		}
		\Else{
			$ \tilde{f}'_{L} := \tilde{f}'_{temp} $\;
			$ \alpha_{L} = \alpha_{temp} $\label{alg2_falsi}
		}
	}
	$ \alpha^{*}_{n} = \alpha_{temp} $\;
	Compute gradient $ \tilde{\boldsymbol{g}}^{*}_{n} $ at $\alpha^{*}_{n}$\;
	\Return{$ \alpha^{*}_{n} $, $ \tilde{\boldsymbol{g}}^{*}_{n} $}
	\caption{Bracketing}
\end{algorithm}

In line~\ref{alg2_init}, we begin with initialization of the lower, $ \alpha_{L} $, and upper, $ \alpha_{U} $, bounds, and their respective directional derivatives, $ \tilde{f}'_{L} $ and $ \tilde{f}'_{U} $. In line~\ref{alg2_grow}, the interpolation interval grows by doubling $ \alpha_{U} $  which is directly followed by $ \alpha_{L} $, until it reaches $\alpha_{max}$ or $ \tilde{f}'_{U} \geq c\tilde{f}'_{0,n} $ is satisfied. In line~\ref{alg2_interp}, the size of the interpolation interval is reduced by consecutively performing linear interpolation until $ \tilde{f}'_{U} \leq -c\tilde{f}_{0,n} $ is satisfied. The rest of conditions in line~\ref{alg2_interp} ensures that linear interpolation steps with the Regula-Falsi method in lines~\ref{alg2_regula}-\ref{alg2_falsi} do not cause any numerical instabilities. The second term, $\tilde{f}'_{U}\tilde{f}'_{L}<0$, ensures that the sign of the two directional derivatives are opposite to each other, and the last term, $|\tilde{f}'_{U}\tilde{f}'_{L}|<\epsilon$, provides the denominator of line~\ref{alg2_regula} to be non-zero.

\begin{figure}
	\centering
	\includegraphics[width=0.4\linewidth]{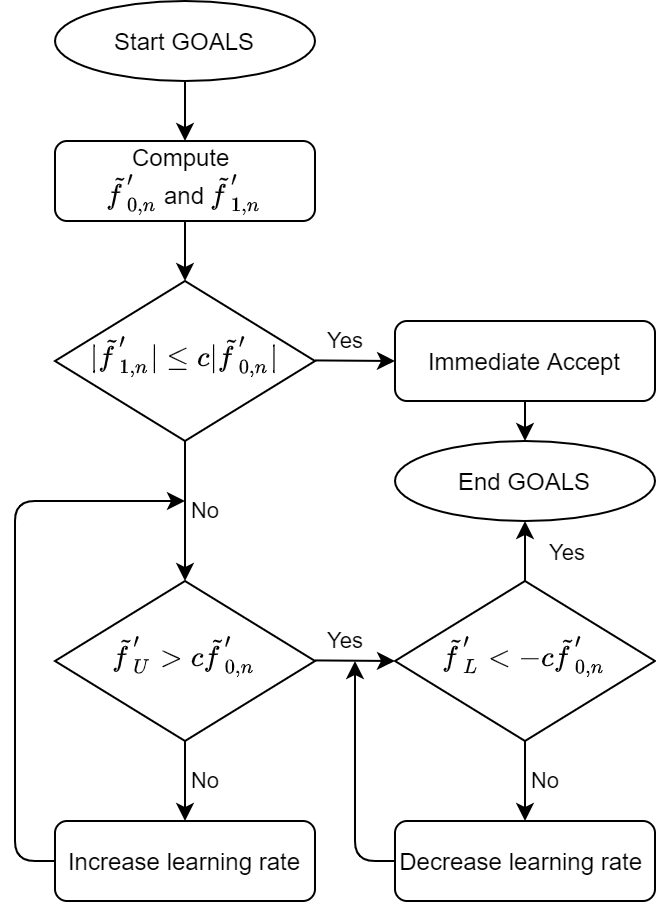}
	\caption{The flowchart of the GOALS line search strategy}
	\label{flowchart2}
\end{figure}

The Wolfe curvature conditions are divided into two sections, as shown in lines~\ref{alg2_grow} and \ref{alg2_interp}, to prevent this algorithm from searching infinitely for points that do not satisfy the Wolfe curvature conditions.  Due to the stochastic nature of dynamic MBSS loss functions, it is not guaranteed that continually reducing learning rates would find a negative directional derivative with less magnitude than the directional derivative at the origin, $ \tilde{f}'_{n,0} $. This implies that we can not assure that the first condition in line~\ref{alg2_grow} is still met after the second condition in line~\ref{alg2_interp} is satisfied. The risk associated with undershooting is lower than that of overshooting because overshooting may cause divergence in training, while undershooting, in the worst case, causes slower training. The flowchart of GOALS is shown in Figure~\ref{flowchart2}. 

\subsection{Proof of convergence}\label{proof}
Let us assume that the full-batch loss function, $ \mathcal{L}(\boldsymbol{x}) $, is a smooth coercive function of a weight vector, $ \boldsymbol{x} \in \mathbb{R}^{p} $, so that we can replace $ \mathcal{L}(\boldsymbol{x})  $ with a Lyapunov function, $ \Gamma(\boldsymbol{x}) $ \citep{lyapunov1992general}. 
The Lyapunov's global stability theorem states that a Lyapunov function, $ \Gamma(\boldsymbol{x}) $, results in $ \lim\limits_{n \rightarrow \infty}\boldsymbol{x}_{n}=0 $, $\forall\enspace \boldsymbol{x}_{n} \in \mathbb{R}^{p}$ under the following conditions:
\begin{enumerate}
	\item Positivity: $ \Gamma(\boldsymbol{0}) = 0 $ and $ \Gamma(\boldsymbol{x})>0 $, $ \forall\enspace\boldsymbol{x} \neq \boldsymbol{0} $
	\item Coercive: $ \lim\limits_{\boldsymbol{x}\rightarrow\infty}\Gamma(\boldsymbol{x}) = \infty$
	\item Strict descent: $ \Gamma(\mathcal{D}(\boldsymbol{x}))<\Gamma(\boldsymbol{x}) $, $ \forall\enspace\boldsymbol{x}\neq\boldsymbol{0}  $
\end{enumerate}
where $ \mathcal{D}(\boldsymbol{x}) $ is a weight update function given by
\begin{equation}\label{key}
	\boldsymbol{x}_{n+1} := \mathcal{D}(\boldsymbol{x}_{n}); \quad  \mathcal{D}:\mathbb{R}^{p} \rightarrow \mathbb{R}^{p}.
\end{equation}
It is proven by \cite{wilke2013gradient} that locating an NN-GPP along a strictly descending direction, $ \boldsymbol{d}_{n} $, is equivalent to minimizing along $ \boldsymbol{d}_{n}  $ when $ \mathcal{L}(\boldsymbol{x}_{n}+ \alpha\boldsymbol{d}_{n}) $ is a smooth function. Therefore, locating NN-GPPs along descent directions in consecutive iterations of a training algorithm behaves like  $ \mathcal{D}(\boldsymbol{x})$. Similarly, for loss functions resulting from dynamic MBSS, $ \tilde{\mathcal{L}} $, we assume as point-wise discontinuous coercive.

Hence, the Lyapunov's global stability theorem is relaxed for the expected Lyapunov function, $ E[\Gamma(\boldsymbol{x})] $, where:
\begin{enumerate}
	\item Expected positivity: $ E[\Gamma(\boldsymbol{0})] = 0$ and $ E[\Gamma(\boldsymbol{x})]>0 $, $ \forall\enspace\boldsymbol{x} \neq \boldsymbol{0} $
	\item Expected coercive: $ \lim\limits_{\boldsymbol{x}\rightarrow\infty}E[(\Gamma(\boldsymbol{x})] = \infty$
	\item Expected strict descent: $ E[\Gamma(\mathcal{D}(\boldsymbol{x}))]<E[\Gamma(\boldsymbol{x})] $, $ \forall\enspace\boldsymbol{x}\neq\boldsymbol{0}  $
\end{enumerate}
Subsequently, consecutively searching for SNN-GPPs along descent directions makes the training algorithm behave like $ \mathcal{D}(\boldsymbol{x}) $, which tends towards a ball, $B_{\epsilon}$:
\begin{equation}\label{key}
	\lim\limits_{\boldsymbol{n}\rightarrow\infty}\boldsymbol{x}_{n} =  \{\boldsymbol{q}| \|\boldsymbol{q}-\boldsymbol{x}^{*}\|<\epsilon\} \in B _{\epsilon}
\end{equation}
where $ B_{\epsilon} $ is a ball function with the radius of $ \epsilon $, with the true optimum, $ \boldsymbol{x}^{*} $, located at its center.
Since our bracketing strategy searches for SNN-GPPs with weights, $ \boldsymbol{x}_{n} \in \mathcal{B}_{\epsilon}$, respectively, along a strictly descending direction, $ \boldsymbol{d}_{n} $,
\begin{equation}\label{key}
	|E[\Gamma(\boldsymbol{x}_{n+1})]-E[\Gamma(\boldsymbol{x}^{*})]| <|E[\Gamma(\boldsymbol{x}_{n})]-E[\Gamma(\boldsymbol{x}^{*})]|,
\end{equation}
and weights outside the ball, $ \boldsymbol{x}_{n} \in \mathcal{B}'_{\epsilon} $, would eventually be inside the ball,  $ \boldsymbol{x}_{n} \in \mathcal{B}_{\epsilon} $, as $ n\rightarrow\infty $.

\section{Numerical study design}\label{numerical_study}
We structured two sets of numerical studies to investigate the performance of our proposed algorithm, GOALS. 
First, we compared GOALS against the fixed learning rates and the vanilla GOS by applying them to popular optimizers such as SGD, \textsc{RMSProp}, and \textsc{Adam}. 
This study aims to show that GOALS is capable of adjusting the learning rates of the various optimizers. 
Second, we examine the performance of GOALS against several other learning rate update strategies, including vanilla GOS, GOLS-I, fixed learning rate methods, and cosine annealing \citep{loshchilov2016sgdr} when the batch sizes change for SGD directions. 
This  compares the performance and relative robustness, $ R $ in \eqref{rr}, of GOALS to the other learning rate strategies.
At the end of each numerical study, we quantify the performance of each strategy by measuring the relative robustness, $ R $, rather than the best performance for both training and testing accuracies.

In this section, after we present the multiple settings of parameter settings selected for GOALS in the numerical study, the details of the two numerical studies are introduced.

\subsection{Parameter settings of GOALS}
GOALS requires three parameters to be selected: 1) the initial learning rate, $ \alpha_{0,1} $, 2) the curvature parameter, $ c $, and 3) a decision on whether we want to use the last learning rate for the following initial learning rate, $ \alpha_{0,n} = \alpha^{*}_{0,n-1} $. The opposite of this would be to reset the next initial learning rate to the default initial learning rate, $ \alpha_{0,n} = \alpha_{0,1} $.

Table~\ref{settings} lists the four different settings of parameters for GOALS. While all four settings keep the curvature parameter, $ c $, identical and large, the initial learning rates, $ \alpha_{0,1} $, are either the recommended learning rates, $ \gamma $, for the selected optimizer or the inverse of the L-2 norm of the search direction, $ 1/\|\boldsymbol{d}_{n}\| $. We use GOALS-1,2,3 for the first numerical study and GOALS-1,2,3,4 for the second numerical study. We omit  GOALS-4 for the first numerical study because the loss function from deep network architectures is highly non-linear. Hence, constructing multiple quadratic approximations causes significant approximation errors and many gradient evaluations when taking an exponentially growing initial guess such as $1/ \|\boldsymbol{d}_{n}\| $.

\begin{table}[h!]
	\begin{center}
		\scalebox{0.8}{
			\begin{tabular}{ccccc}
				\hline
				   Algorithms                      & Settings &       $\alpha_{0,1}$       & $\alpha_{0,n} = \alpha^{*}_{0,n-1}$ & $c$ \\ \hline
				\multirow{4}{4em}{GOALS} & GOALS-1  &          $\gamma$          &                 No                  & 0.9 \\
				                         & GOALS-2  &          $\gamma$          &                 Yes                 & 0.9 \\
				                         & GOALS-3  & $1/\|\boldsymbol{d}_{n}\|$ &                 Yes                 & 0.9 \\
				                         & GOALS-4  & $1/\|\boldsymbol{d}_{n}\|$ &                 No                  & 0.9 \\ \hline
				          GOS            &    -     & $1/\|\boldsymbol{d}_{n}\|$ &                 No                  &  -  \\ \hline
			\end{tabular}}
	\end{center}
	\caption{Comparison between the settings of vanilla GOS and GOALS that are tested in this paper. We choose the initial learning rates for the first iteration, $\alpha_{0,1}$, the curvature parameter, $c$, and decide whether we want to use the final learning rate as the next initial learning rate, $\alpha_{0,n} = \alpha^{*}_{0,n-1}$.}
	\label{settings}
\end{table}

\subsection{Numerical study 1 setup }
For the first numerical study, we investigate the learning rate adaption of the proposed line search algorithm for different architectures and optimizers. 
It is compared against GOS and fixed learning rates for various optimizers' search directions: SGD, \textsc{RMSProp}, and \textsc{Adam}. 
We chose ResNet-18 \citep{he2016deep} and EfficientNet-B0 \citep{tan2019efficientnet}, which are implemented by \cite{Liu2020}, for the test DNN models and the CIFAR-10  dataset \citep{Krizhevsky2009} for this experiment. 
The details of the dataset are shown in Table~\ref{datasets}. The chosen mini-batch size,  $|\mathcal{B}|$, for this numerical study is 128.

The following learning rate strategies were trained for 350 epochs, repeated five times: fixed learning rate, GOS, GOALS-1, GOALS-2, GOALS-3. The fixed learning rates, $ \gamma $, for SGD, \textsc{RMSprop}, and \textsc{Adam} are 0.01, 0.01, and 0.001, respectively, which are the default values provided by PyTorch \cite{NEURIPS2019_9015} and TensorFlow \cite{tensorflow2015-whitepaper}. \revOne{Note that because we adopt dynamic MBSS, a different mini-batch for every function evaluation, for the experiments, the fixed number of epoch also means the fixed number of gradient computations in training. Hence, some strategies may have fewer search directions when more gradient evaluations are required for each search direction or iteration.}

\begin{table}[h!]
	\begin{center}
		\scalebox{0.8}{
			\begin{tabular}{ccccc}
				\hline
				Datasets             & Classes &  Input sizes  & Training samples &  Test samples   \\ \hline
				MNIST \citep{lecun1998gradient}      &   10    & $28\times 28$ & $6\times10^{4}$  & $1\times10^{4}$ \\
				CIFAR-10 \citep{Krizhevsky2009} &   10    & $32\times 32$ & $5\times10^{4}$  & $1\times10^{4}$ \\ \hline
		\end{tabular}}
	\end{center}
	\caption{Descriptions of datasets used in the numerical study}
	\label{datasets}
\end{table}

\subsection{Numerical study 2 setup}
\revOne{For the second numerical study, we first examine the performance of GOALS with different hyperparameter settings: GOALS-1, 2, 3, and 4 to choose the most robust one based on $ R $.  
Next, we test the chosen one against various learning rate strategies such as GOS, GOLS-I, fixed learning rate methods, cosine annealing \citep{loshchilov2016sgdr} different mini-batch sizes. 
We restrict ourselves to SGD directions on shallower neural network architecture.}

We used similar neural network training problems to those proposed by \cite{mahsereci2017probabilistic}, namely training on a fully-connected feedforward neural network problem, N-II. 
This network's architecture involves fully connected layers with three hidden layers, $n_{input}-1000-500-250-n_{output}$. 
It contains the $ \tanh $ activation functions, mean square loss, and Xavier initialization \citep{glorot2010understanding}. 
The dataset used for the problem is MNIST in Table~\ref{datasets}.

The descriptions of the learning rate strategies compared against our algorithms are listed as follows:
\begin{enumerate}
	\item Fixed learning rates: we test five sets of fixed learning rates, $\alpha = 10^{-3}$, $10^{-2}$, $10^{-1}$, $10^{0}$ and, $10^{1}$;
	\item The Cosine annealing scheduler with warm restarts \citep{loshchilov2016sgdr}: starting learning rates used are $ \alpha= 10^{-1} $ and $ 10^{0} $, the initial restart period, the multiplying factor is chosen to be $ T_{0} = 1$ epoch, and $ T_{mult} = 2$;
	\item The Gradient-only line search that is Inexact (GOLS-I) \citep{kafka2019gradient};
	\item The vanilla gradient-only surrogate/approximation (GOS) line search \citep{chae2019empirical}.
	\item The gradient-only approximation line search (GOALS) allows various parameter settings. For the comparison, we choose GOALS-4 in Table~\ref{settings}.
\end{enumerate}

The training is limited in the number of directional derivative computations per training run, and the limit is $4\times 10^{4}$. The mini-batch sizes chosen were $ |\mathcal{B}| \in \{10,100,200,1000\} $. As mentioned earlier, we select the SGD direction as the search directions, computed using the same mini-batch size, $ |\mathcal{B}| $. For each setting, we take ten runs for generating results.

\section{Results of numerical study}
\subsection{Results of numerical study 1}
\begin{figure}[h!]
	\begin{subfigure}[b]{1\textwidth}
		\centering
		\includegraphics[width=\textwidth]{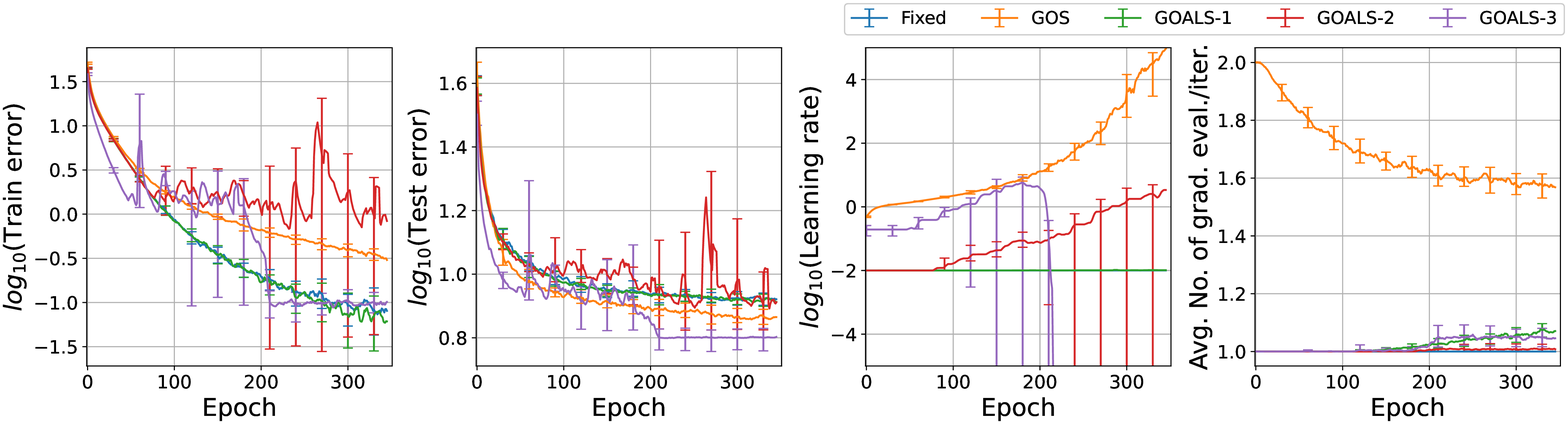}
		\caption{SGD}
	\end{subfigure}
	\begin{subfigure}[b]{1\textwidth}
		\centering
		\includegraphics[width=\textwidth]{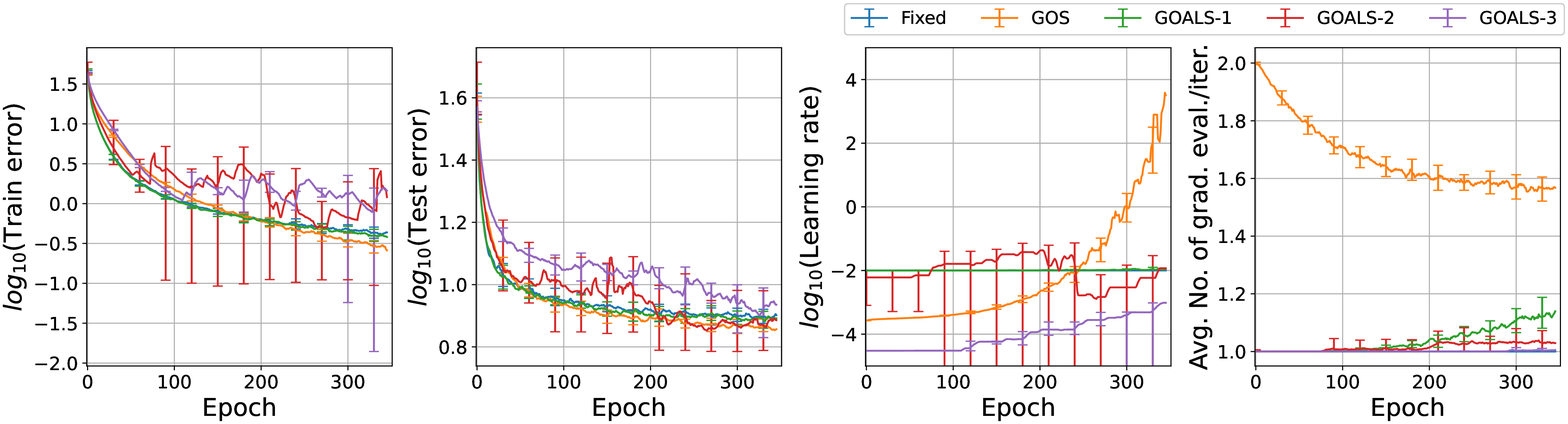}
		\caption{\textsc{RMSprop}}
	\end{subfigure}
	\begin{subfigure}[b]{1\textwidth}
		\centering
		\includegraphics[width=\textwidth]{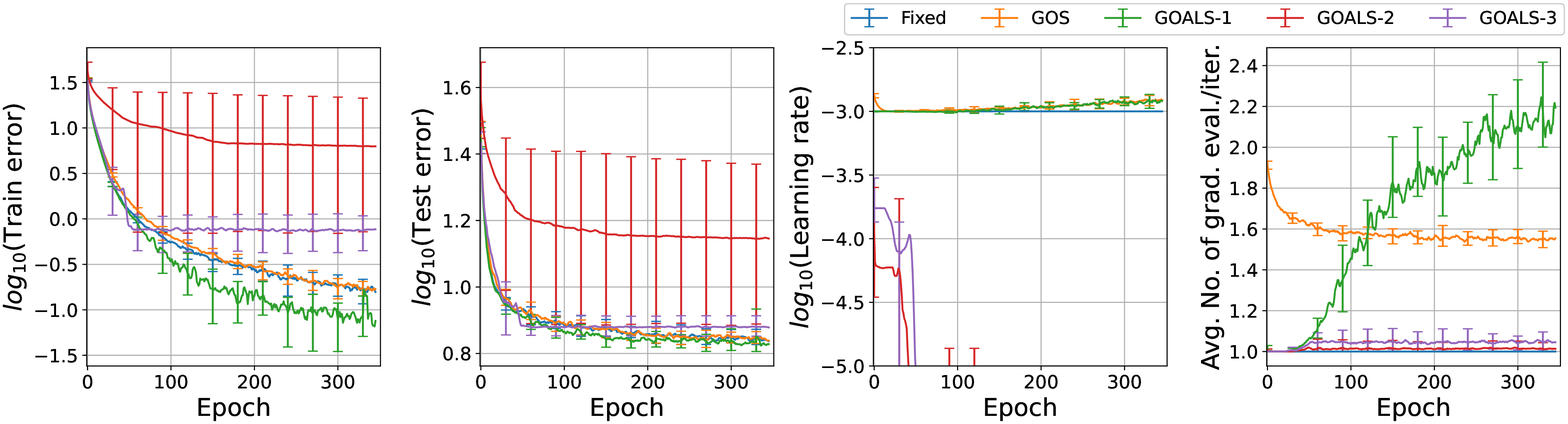}
		\caption{\textsc{Adam}}
	\end{subfigure}
	\caption{Comparisons of the performances of (a)~SGD, (b)~\textsc{RMSprop}, (c)~\textsc{Adam} between with and without GOALS applied, tested on ResNet-18 for the CIFAR-10 dataset, the results are averaged over five runs and smoothened out with moving average over five epochs.}
	\label{resnet18}
\end{figure}

\begin{figure}[h!]
	\begin{subfigure}[b]{1\textwidth}
		\centering
		\includegraphics[width=\textwidth]{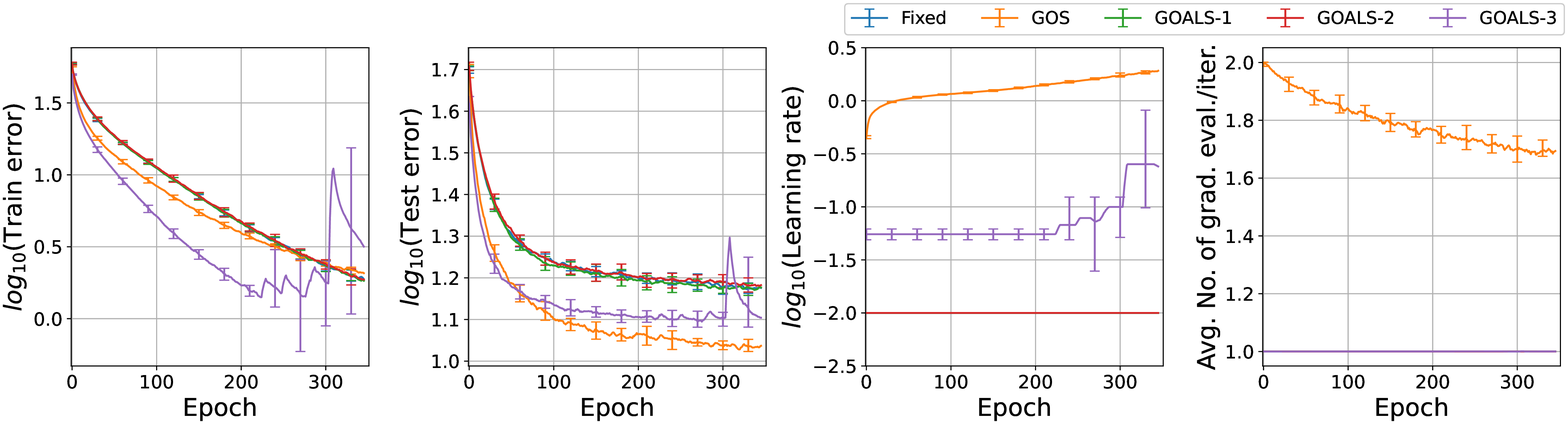}
		\caption{SGD, $\alpha= 0.01 $}
	\end{subfigure}
	\begin{subfigure}[b]{1\textwidth}
		\centering
		\includegraphics[width=\textwidth]{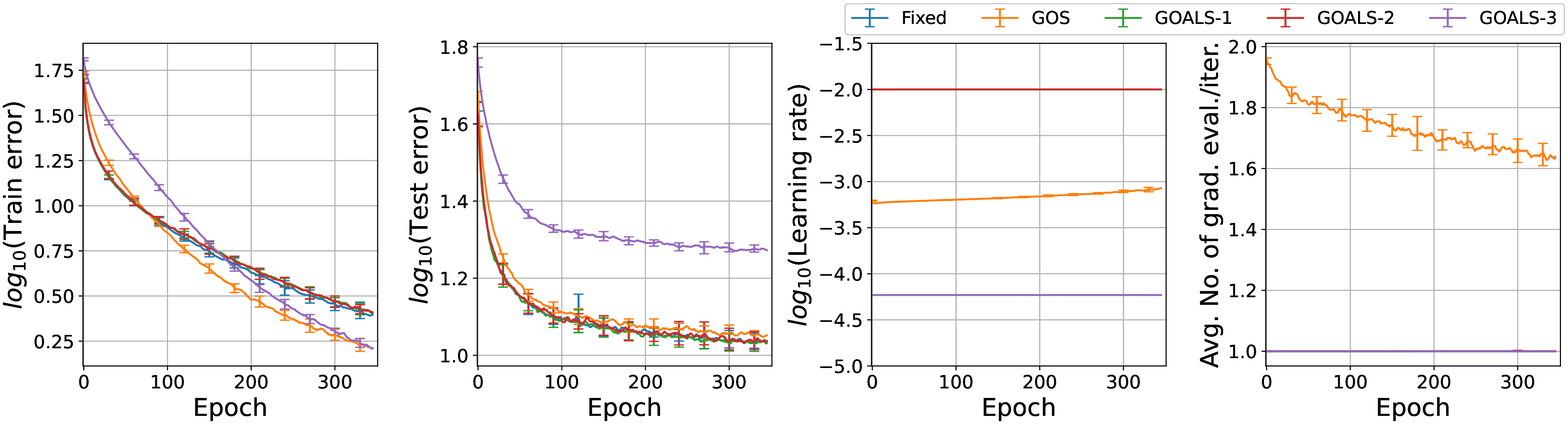}
		\caption{\textsc{RMSprop}}
	\end{subfigure}
	\begin{subfigure}[b]{1\textwidth}
		\centering
		\includegraphics[width=\textwidth]{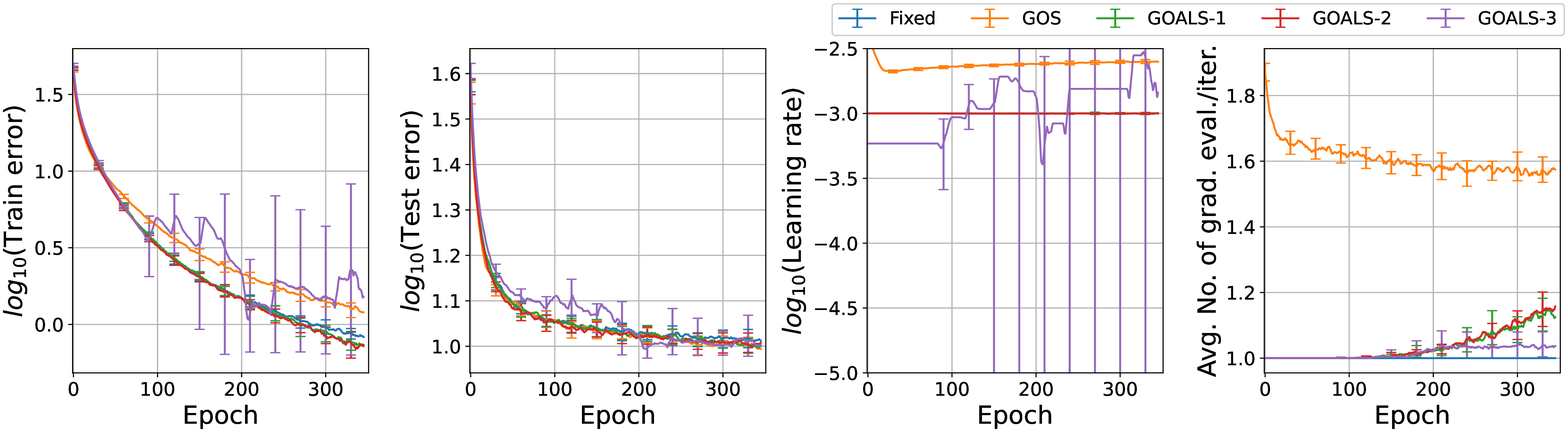}
		\caption{\textsc{Adam}}
	\end{subfigure}
	\caption{Comparisons of the performances of (a)~SGD, (b)~\textsc{RMSprop}, (c)~\textsc{Adam} between with and without GOALS applied, tested on EfficientNet-B0 for the CIFAR-10 dataset, the results are averaged over five runs and smoothened out with moving average over five epochs.}
	\label{efficientB0}
\end{figure}

Figures~\ref{resnet18} and \ref{efficientB0} show the results for SGD, \textsc{RMSprop} and \textsc{Adam}. 
For each algorithm, the 5-step moving average values of the training errors, test errors, learning rates, and the number of gradient computations are plotted along 350 epochs using error bars on a log-10 scale. 
The lower errors and upper errors represent the minimum and the maximum errors of the five runs. 
Note that every function evaluation of GOALS requires a new mini-batch. 
The larger the number of function evaluations computed per iteration, the fewer search direction updates per epoch. 

Since GOALS is a conservative algorithm with convergence proof instead of GOS, which is not conservative, we expect GOALS to converge less aggressively than GOS but with more confidence. A common phenomenon observed in most of the results in Figures~\ref{resnet18} and \ref{efficientB0} is that the average learning rates of both GOS and GOALS increase as the epoch increases. This happens because the directional derivative at the origin, $\tilde{f}'_{0,n}$, decreases over time. This means that the average number of gradient computations for GOALS may increase over epochs to satisfy the curvature condition on the flatter domains of the functions. On the other hand, that for GOS may decrease since the chance of observing a directional derivative at $\alpha_{1,n}$, $\tilde{f}'_{1,n}$, that is less than the initial directional derivative, $\tilde{f}'_{0,n}$, grows. In this case, it accepts $\alpha_{1,n}$ as the final learning rate, $\alpha^{*}_{n}$.

The ResNet results shown in Figure~\ref{resnet18} indicate that the initial convergence rate in GOS's training is slightly lower than for the fixed learning rates and GOALS. This is because it does not extend the learning rate to be inside the ball. Figure~\ref{resnet18}(a) shows that although the training error of GOS is relatively high, its test error is one of the lowest. The performance of GOALS-1 is similar to that of the fixed learning rate since the initial learning rate, $\alpha_{0,n} = \gamma$, happens to satisfy the Wolfe condition most of the time. Hence, note that there is a slightly increasing number of gradient evaluations only close to the training's end. GOALS-2 shows a large variance in the performance since using previous learning rates may cause a large model error as the previous learning rate might be outside the ball. Hence, it is more challenging to find the SNN-GPPs for multimodal distributions of sign changes. The lower limit of GOALS learning rates is softly bounded since we satisfy the lower curvature condition before satisfying the upper independently.

Consequently, when the model error is large, it may allow learning rates that are numerically zero when using the previous learning rates. For the same reason, we also observe the phenomenon of diminishing learning rate in GOALS-3. However, it starts with larger learning rates and a steeper convergence rate in training, and it also shows the lowest test error for SGD. 

Figure~\ref{resnet18}(b) shows that the ResNet results for \textsc{RMSprop} show that GOS has the lowest training error, while both GOALS-2 and GOALS-3 show large fluctuations in learning rates. GOALS-1 and the fixed learning rate perform similarly since the recommended learning rate approximates SNN-GPPs well. 

The ResNet results for \textsc{Adam} show that while both GOALS-2 and GOALS-3 again perform poorly, both GOS and the fixed learning rate perform similarly. Note that GOALS-1, whose learning rates are very similar to GOS, shows the best training and testing performance. Also, note that the average number of gradient computations for GOALS-1 increases as the epoch increases, and it shows the largest values among the optimizers.

The EfficientNet-B0 results shown in Figure~\ref{efficientB0} indicate that SGD and \textsc{RMSprop} are mostly unaffected by GOALS-1 and GOALS-2. Their default learning rate mostly satisfies the curvature condition. However, \textsc{Adam} with GOALS affected the learning rates close to the end of training with slightly lower training and testing errors. GOALS-3 for SGD, shown in Figure~\ref{efficientB0}(a), has a high initial convergence rate. However, the large variance in learning rates resulted in fluctuations in the training error. Although GOALS-3 shows the lowest training error, while GOS shows the lowest testing error. Figure~\ref{efficientB0}(c) shows the average numbers of gradient computation for both GOALS-1 and GOALS-2 increases for \textsc{Adam}, unlike SGD and \textsc{RMSprop}. This results in improved train accuracies compared to the fixed learning rate. 

We listed the top average training and testing accuracies of the optimizers with GOS and GOALS with various initial point settings in Table~\ref{performance}. The results tell us that the SGD search direction works best with the vanilla GOS in terms of generalization shown by the test accuracy. The next lesson is that when using the previous learning rate as the next initial guess, $\alpha_{0,n} = \alpha^{*}_{n-1}$, such as GOALS-2 and GOALS-3, may result in severe quadratic model errors. Hence, we do not recommend this approach. Instead, when the initial learning rate is the recommended learning rate, $\gamma$, for the chosen search direction such as GOALS-1, provides stable and improved performance in most cases, especially, \textsc{Adam} which has the momentum effects.

\begin{table}
	\begin{center}
		\scalebox{0.9}{
			\begin{tabular}{lllllll}
				\hline
				Models                            & Optimizers                          & Methods & Train acc. [\%]        & Diff., $ \psi_{y,h,o} $ & Test acc. [\%]         & Diff., $ \psi_{y,h,o} $ \\ \hline
				\multirow{20}{7em}{ResNet-18}     & \multirow{5}{4em}{SGD}              & Fixed   & 99.94                  & 0.02                    & 91.88                  & 1.9                     \\
				                                  &                                     & GOS     & 99.72 (-0.22)          & 0.24                    & 92.93 (+1.05)          & 0.85                    \\
				                                  &                                     & GOALS-1 & \textbf{99.96 (+0.02)} & 0                       & 91.96 (+0.08)          & 1.82                    \\
				                                  &                                     & GOALS-2 & 99.43 (-0.51)          & 0.53                    & 92.33 (+0.45)          & 1.45                    \\
				                                  &                                     & GOALS-3 & 99.93 (-0.01)          & 0.03                    & \textbf{93.78 (+1.90)} & 0                       \\ \cline{2-7}
				                                  & \multirow{5}{4em}{\textsc{RMSprop}} & Fixed   & 99.64                  & 0.14                    & 92.37                  & 0.65                    \\
				                                  &                                     & GOS     & \textbf{99.78 (+0.14)} & 0                       & \textbf{93.02 (+0.65)} & 0                       \\
				                                  &                                     & GOALS-1 & 99.65 (+0.01)          & 0.13                    & 92.59 (+0.12)          & 0.43                    \\
				                                  &                                     & GOALS-2 & 99.56 (-0.08)          & 0.22                    & \textbf{93.02 (+0.65)} & 0                       \\
				                                  &                                     & GOALS-3 & 99.25 (-0.39)          & 0.53                    & 92.24 (-0.13)          & 0.78                    \\ \cline{2-7}
				                                  & \multirow{5}{4em}{\textsc{Adam}}    & Fixed   & 99.87                  & 0.09                    & 93.3                   & 0.28                    \\
				                                  &                                     & GOS     & 99.86 (-0.01)          & 0.1                     & 93.23 (-0.07)          & 0.35                    \\
				                                  &                                     & GOALS-1 & \textbf{99.96 (+0.09)} & 0                       & \textbf{93.58 (+0.28)} & 0                       \\
				                                  &                                     & GOALS-2 & 93.79 (-6.08)          & 6.17                    & 86.08 (-7.22)          & 7.5                     \\
				                                  &                                     & GOALS-3 & 99.31 (-0.56)          & 0.65                    & 92.53 (-0.77)          & 1.05                    \\ \cline{2-7}
				                                  & \multirow{5}{4em}{Sum, $ R_{y,h} $} & Fixed   & \multicolumn{1}{c}{-}                      & 0.25                    & \multicolumn{1}{c}{-}                      & 2.83                    \\
				                                  &                                     & GOS     & \multicolumn{1}{c}{-}                      & 0.34                    & \multicolumn{1}{c}{-}                      & \textbf{1.2}            \\
				                                  &                                     & GOALS-1 & \multicolumn{1}{c}{-}                      & \textbf{0.13}           & \multicolumn{1}{c}{-}                      & 2.25                    \\
				                                  &                                     & GOALS-2 & \multicolumn{1}{c}{-}                      & 6.92                    & \multicolumn{1}{c}{-}                      & 8.95                    \\
				                                  &                                     & GOALS-3 & \multicolumn{1}{c}{-}                      & 1.21                    & \multicolumn{1}{c}{-}                      & 1.83                    \\ \hline
				\multirow{15}{*}{EfficientNet-B0} & \multirow{5}{4em}{SGD}              & Fixed   & 98.18                  & 0.46                    & 85.37                  & 4.07                    \\
				                                  &                                     & GOS     & 97.94 (-0.24)          & 0.7                     & \textbf{89.44 (+4.07)} & 0                       \\
				                                  &                                     & GOALS-1 & 98.25 (+0.07)          & 0.39                    & 85.39 (+0.02)          & 4.05                    \\
				                                  &                                     & GOALS-2 & 98.15 (-0.03)          & 0.49                    & 84.97 (-0.4)           & 4.47                    \\
				                                  &                                     & GOALS-3 & \textbf{98.64 (+0.46)} & 0                       & 87.8 (+1.43)           & 1.64                    \\ \cline{2-7}
				                                  & \multirow{5}{4em}{\textsc{RMSprop}} & Fixed   & 97.6                   & 0.83                    & 89.44                  & 0.11                    \\
				                                  &                                     & GOS     & \textbf{98.43 (+0.83)} & 0                       & 89.0 (-0.44)           & 0.55                    \\
				                                  &                                     & GOALS-1 & 97.51 (-0.09)          & 0.92                    & \textbf{89.55 (+0.11)} & 0                       \\
				                                  &                                     & GOALS-2 & 97.46 (-0.14)          & 0.97                    & 89.42 (-0.02)          & 0.13                    \\
				                                  &                                     & GOALS-3 & 98.4 (+0.8)            & 0.03                    & 81.51 (-7.93)          & 8.04                    \\ \cline{2-7}
				                                  & \multirow{5}{4em}{\textsc{Adam}}    & Fixed   & 99.21                  & 0.15                    & 89.96                  & 0.34                    \\
				                                  &                                     & GOS     & 98.85 (-0.36)          & 0.51                    & 90.22 (+0.26)          & 0.08                    \\
				                                  &                                     & GOALS-1 & 99.34 (+0.13)          & 0.02                    & 90.22 (+0.26)          & 0.08                    \\
				                                  &                                     & GOALS-2 & \textbf{99.36 (+0.15)} & 0                       & 90.13 (+0.17)          & 0.17                    \\
				                                  &                                     & GOALS-3 & 99.0 (-0.21)           & 0.36                    & \textbf{90.3 (+0.34)}  & 0                       \\ \cline{2-7}
				                                  & \multirow{5}{4em}{Sum, $ R_{y,h} $} & Fixed   & \multicolumn{1}{c}{-}                      & 1.44                    & \multicolumn{1}{c}{-}                      & 4.52                    \\
				                                  &                                     & GOS     & \multicolumn{1}{c}{-}                      & 1.21                    & \multicolumn{1}{c}{-}                      & \textbf{0.63}           \\
				                                  &                                     & GOALS-1 & \multicolumn{1}{c}{-}                      & 1.33                    & \multicolumn{1}{c}{-}                     & 4.13                    \\
				                                  &                                     & GOALS-2 & \multicolumn{1}{c}{-}                      & 1.46                    & \multicolumn{1}{c}{-}                      & 4.77                    \\
				                                  &                                     & GOALS-3 & \multicolumn{1}{c}{-}                      & \textbf{0.39}           & \multicolumn{1}{c}{-}                      & 9.68                    \\ \hline
				\multirow{5}{*}{Overall, $ R_{y} $} &                                     & Fixed   & \multicolumn{1}{c}{-}                      & 1.69                    &                      \multicolumn{1}{c}{-}  & 7.35                    \\
				                                  &                                     & GOS     & \multicolumn{1}{c}{-}                      & 1.55                    &                   \multicolumn{1}{c}{-}     & \textbf{1.83}           \\
				                                  &                                     & GOALS-1 & \multicolumn{1}{c}{-}                      & \textbf{1.46}           &                     \multicolumn{1}{c}{-}   & 6.38                    \\
				                                  &                                     & GOALS-2 & \multicolumn{1}{c}{-}                      & 8.38                    &                    \multicolumn{1}{c}{-}    & 13.72                   \\
				                                  &                                     & GOALS-3 & \multicolumn{1}{c}{-}                      & 1.6                     &                   \multicolumn{1}{c}{-}    & 11.51                   \\ \hline												  
			\end{tabular}}
	\end{center}
	\caption{\revOne{Top average training and testing accuracies over the five runs tabulated for widely-used optimizers, including SGD, \textsc{RMSprop} and \textsc{Adam}, with vanilla GOS and GOALS with various settings on ResNet-18 and EfficientNet-B0. 
	The differences in performance compared to the fixed learning rate are given inside the brackets.
	It also measures the relative robustness, $ R_{y,h} $, for each problem by computing the difference, $ \psi_{y,h,o} $, between the performance and the best one from the same batch size.
	The overall relative robustness, $ R_{y} $, in the last section sums all the differences measured for each strategy, $ y $ in each problem, $ h $.  
	The highest train, test accuracies, and the lowest robustness measures are indicated in bold.}}
	\label{performance}
\end{table}

\revOne{We motivate the choice of GOALS-1 for DNN by our relative robustness measure. 
The difference between the performance of each strategy and the best performing strategy for each optimizer is measured. 
The relative robustness is then computed as the sum of the difference values. 
Overall, the last section in the table is the overall robustness measure for both ResNet-18 and EfficientNet-B0
Note that GOALS-1 gives the smallest sum of difference for both training and testing errors. 
Hence, it is most robust to our measure. 
It means that although the hyperparameter setting may not be the best performing for a specific neural network, we expect it to fail less to an unseen problem.}

In \ref{comparisons}, we have an additional numerical experiment conducted to test the performance of our proposed line search method, GOALS, on a much shallower, 3-hidden-layer-network problem.

\subsection{Results of numerical study 2} \label{comparisons}
\subsubsection{Performance test for different hyperparameter settings}
\begin{figure}[h!]
	\centering
	\includegraphics[width=\textwidth]{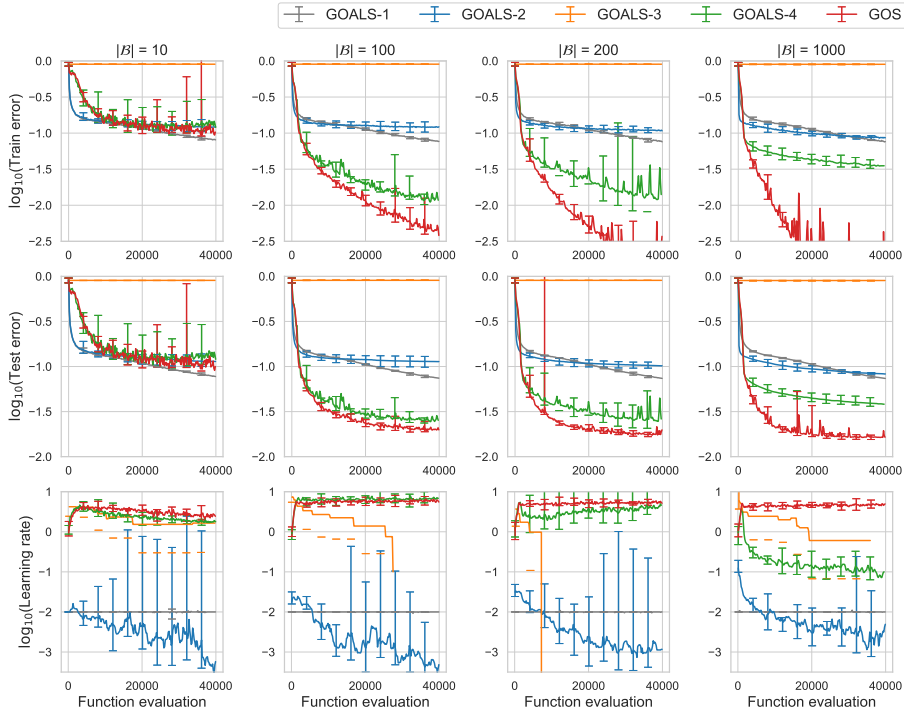}
	\caption{N-II MNIST dataset with batch size, $|\mathcal{B}|$ = 10, 100, 200 and 1000 from left to right for various parameters settings of GOALS which are listed in Table~\ref{settings}. The comparison of training dataset error (the 1st row), test set error (the 2nd row) and learning rate (the 3rd row) on a $ log_{10} $ scale versus the number of function evaluations.}
	\label{n2result2}
\end{figure}

Before we begin to compare the different strategies, we investigate GOALS results with various settings shown in Table~\ref{settings} on the N-II architecture. Figure~\ref{n2result2} shows the training error, testing error, and learning rates for the various hyperparameter settings on the N-II architecture with MNIST. Note that the training and testing errors for MNIST are plotted on the log-10 scale, averaged over ten runs. It is noticed that GOALS-1 made unnoticeable changes in learning rates to the initial learning rates of 0.01, since the recommended learning rates continuously satisfied the curvature condition, resulted in the IAC condition. Hence, it performed almost exactly to the fixed learning rate of 0.01.

On the other hand, GOALS-4, whose setting is similar to GOALS-1 except for the initial guess, outperforms all the rest and the learning rate is indeed close to the vanilla GOS. However, due to the curvature condition, we note that the learning rates are lower than GOS, especially for larger batch sizes. Hence, we observe the best performance of GOALS-4 when the batch size is 100.

\revOne{Table~\ref{mnist_table2} shows the top average training and testing accuracies for each hyperparameter setting, and the overall robustness measures for them on N-II are listed in the last row. We find that the most robust strategy is GOS, in which the measures are the smallest, followed by GOALS-4. Hence, we choose GOALS-4 for the further performance comparison on the shallower network, N-II.}

\begin{table}[h!]
	\begin{center}
		\scalebox{0.9}{
			\begin{tabular}{clllll}
				\hline
				\multicolumn{1}{l}{Batch size, $ |\mathcal{B}| $} & Methods & Train acc. & Diff., $ \psi_{y,|\mathcal{B}|} $  & Test acc. & Diff., $ \psi_{y,|\mathcal{B}|} $ \\ \hline
				\multirow{5}{*}{10}       & GOALS-1 & \textbf{91.95}      & 0      & \textbf{92.26}     & 0     \\
				& GOALS-2 & 88.05      & 3.9    & 88.63     & 3.63  \\
				& GOALS-3 & 10.05      & 81.9   & 10.02     & 82.24 \\
				& GOALS-4 & 89.4       & 2.55   & 89.55     & 2.71  \\
				& GOS     & 90.62      & 1.33   & 90.93     & 1.33  \\ \hline
				\multirow{5}{*}{100}      & GOALS-1 & 92.33      & 7.29   & 92.57     & 5.49  \\
				& GOALS-2 & 88.05      & 11.57  & 88.66     & 9.4   \\
				& GOALS-3 & 9.95       & 89.67  & 9.93      & 88.13 \\
				& GOALS-4 & 98.89      & 0.73   & 97.55     & 0.51  \\
				& GOS     & \textbf{99.62}      & 0      & \textbf{98.06}     & 0     \\ \hline
				\multirow{5}{*}{200}      & GOALS-1 & 92.43      & 7.44   & 92.62     & 5.64  \\
				& GOALS-2 & 89.23      & 10.64  & 89.83     & 8.43  \\
				& GOALS-3 & 9.98       & 89.89  & 9.78      & 88.48 \\
				& GOALS-4 & 98.8       & 1.07   & 97.53     & 0.73  \\
				& GOS     & \textbf{99.87}      & 0      & \textbf{98.26}     & 0     \\ \hline
				\multirow{5}{*}{1000}      & GOALS-1 & 92.44      & 7.49   & 92.62     & 5.75  \\
				& GOALS-2 & 91.49      & 8.44   & 91.74     & 6.63  \\
				& GOALS-3 & 10.31      & 89.62  & 10.12     & 88.25 \\
				& GOALS-4 & 96.51      & 3.42   & 96.18     & 2.19  \\
				& GOS     & \textbf{99.93}      & 0      & \textbf{98.37}     & 0     \\ \hline
				\multirow{5}{*}{Overall, $ R_{y} $}     & GOALS-1 &      \multicolumn{1}{c}{-}      & 22.22  &   \multicolumn{1}{c}{-}        & 16.88 \\
				& GOALS-2 &     \multicolumn{1}{c}{-}       & 34.55  &      \multicolumn{1}{c}{-}     & 28.09 \\
				& GOALS-3 &      \multicolumn{1}{c}{-}      & 351.08 &    \multicolumn{1}{c}{-}       & 347.1 \\
				& GOALS-4 &     \multicolumn{1}{c}{-}       & 7.77   &   \multicolumn{1}{c}{-}        & 6.14  \\
				& GOS     &       \multicolumn{1}{c}{-}     & \textbf{1.33}   &    \multicolumn{1}{c}{-}       & \textbf{1.33}  \\ \hline
		\end{tabular}}
	\end{center}
	\caption{\revOne{Top average training and testing accuracies over the ten runs for the various GOALS settings, GOALS-1, GOALS-2, GOALS-3, GOALS-4, and GOS on the N-II architecture with different batch sizes, $ |\mathcal{B}| = $ 10, 100, 200, 1000. It also measures the overall relative robustness, $ R_{y} $, by computing the difference, $ \psi_{y,|\mathcal{B}|} $, between the performance and the best one from the different batch sizes. The highest train, test accuracies, and the lowest robustness measures are indicated in bold.}}
	\label{mnist_table2}
\end{table}

\subsubsection{Performance comparison for different strategies}
\begin{figure}[h!]
	\centering
	\includegraphics[width=\textwidth]{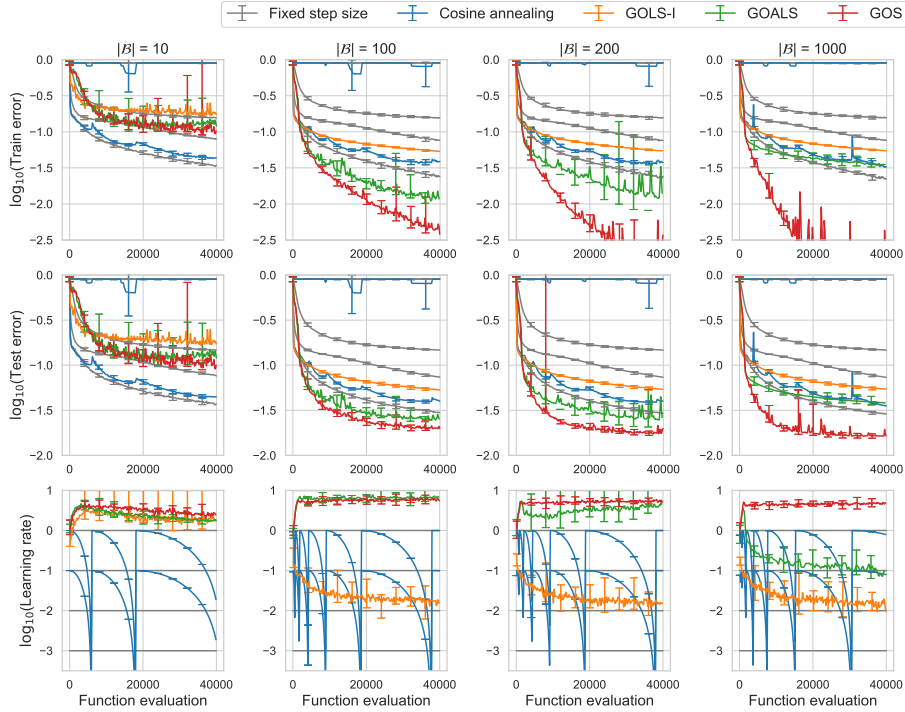}
	\caption{N-II MNIST dataset with batch size, $|\mathcal{B}|$ = 10, 100, 200 and 1000 from left to right for various line search methods: constant learning rates, cosine annealing, GOLS-I, vanilla GOS, and GOALS. The comparison of training dataset error (the 1st row), test set error (the 2nd row) and learning rate (the 3rd row) on a log-10 scale versus the number of function evaluations.}
	\label{n2result1}
\end{figure}

Next, Figure~\ref{n2result1} shows the training error, testing error, and learning rates for the various strategies on the N-II architecture with MNIST. The constant learning rates generally show low variance in error during training. Its error noticeably reduces from $|\mathcal{B}| = 10$ to $|\mathcal{B}| = 100 $ since the SGD direction becomes more representative of the exact (full-batch) SGD direction. The learning rates are independent of any available information when $|\mathcal{B}|$ is small and the information is noisy. Hence, the more straightforward strategies may perform better than more sophisticated ones. 

The best performing strategy for SGD overall is GOS, which is consistent with the observation in Section~\ref{numerical_study} and the reason is that SGD is less sensitive to overshooting. GOS, which is not controlled by convergence theorem, produces relatively larger learning rates. The training performance of GOS noticeably improves as $ |\mathcal{B}| $ increases.

Cosine annealing with warm restarts changes the learning rates over periods of epoch and resets them to the initial learning rates. Hence, we observe slight fluctuations in both training and testing errors. Although this method sweeps through an extensive range of learning rate, the results show that initial learning rate choice significantly affects its performance. 

Both GOALS and GOLS-I try to locate SNN-GPPs. However, GOALS, which uses the quadratic model, outperforms GOLS-I without an approximation model. Both strategies double the learning rates to grow, but GOALS uses quadratic approximations to shrink, while GOLS-I halves the learning rates to shrink. Hence, GOALS may perform worse if the quadratic approximation has a large approximation error due to noisy information.

\revOne{Table~\ref{mnist_table1} shows the top average training and testing accuracies for each hyperparameter setting, and the overall robustness measures for them on N-II are listed in the last row. Since GOALS-4 locates minima using approximation, when the batch size is as small as 10, its robustness is not among the best, including GOS. However, as the batch size gets larger, both GOALS-4 and GOS outperform the most strategies. If we remove the impractically small step size of 10 from the scope, the score of GOALS-4 becomes 3.43, which is the second most robust.}

\begin{table}[h!]
	\begin{center}
		\scalebox{0.9}{
			\begin{tabular}{clllll}
				\hline
				\multicolumn{1}{l}{Batch size, $ |\mathcal{B}|$} & Methods                 & Train acc.            & Diff., $ \psi_{y,|\mathcal{B}|} $ & Test acc.             & Diff., $\psi_{y,|\mathcal{B}|}$ \\ \hline
				              \multirow{10}{*}{10}               & Fixed, $\alpha = 0.001$ & 84.54                 & 12.13                             & 85.3                  & 11.06                           \\
				                                                 & Fixed, $\alpha = 0.01$  & 92.1                  & 4.57                              & 92.4                  & 3.96                            \\
				                                                 & Fixed, $\alpha = 0.1$   & \textbf{96.67}        & 0                                 & \textbf{ 96.36}       & 0                               \\
				                                                 & Fixed, $\alpha = 1$     & 10.09                 & 86.58                             & 10.14                 & 86.22                           \\
				                                                 & Fixed, $\alpha = 10$    & 10.05                 & 86.62                             & 9.74                  & 86.62                           \\
				                                                 & Cosine, $\alpha = 0.1$  & 95.73                 & 0.94                              & 95.55                 & 0.81                            \\
				                                                 & Cosine, $\alpha = 1$    & 36.2                  & 60.47                             & 36.4                  & 59.96                           \\
				                                                 & GOLS-I                  & 83.29                 & 13.38                             & 83.39                 & 12.97                           \\
				                                                 & GOALS-4                 & 89.4                  & 7.27                              & 89.55                 & 6.81                            \\
				                                                 & GOS                     & 90.62                 & 6.05                              & 90.93                 & 5.43                            \\ \hline
				             \multirow{10}{*}{100}               & Fixed, $\alpha = 0.001$ & 84.59                 & 15.03                             & 85.27                 & 12.79                           \\
				                                                 & Fixed, $\alpha = 0.01$  & 92.41                 & 7.21                              & 92.61                 & 5.45                            \\
				                                                 & Fixed, $\alpha = 0.1$   & 97.64                 & 1.98                              & 97.02                 & 1.04                            \\
				                                                 & Fixed, $\alpha = 1$     & 10.05                 & 89.57                             & 9.95                  & 88.11                           \\
				                                                 & Fixed, $\alpha = 10$    & 10.03                 & 89.59                             & 9.86                  & 88.2                            \\
				                                                 & Cosine, $\alpha = 0.1$  & 96.35                 & 3.27                              & 96.03                 & 2.03                            \\
				                                                 & Cosine, $\alpha = 1$    & 36.68                 & 62.94                             & 36.84                 & 61.22                           \\
				                                                 & GOLS-I                  & 94.66                 & 4.96                              & 94.68                 & 3.38                            \\
				                                                 & GOALS-4                 & 98.89                 & 0.73                              & 97.55                 & 0.51                            \\
				                                                 & GOS                     & \textbf{99.62}        & 0                                 & \textbf{98.06}        & 0                               \\ \hline
				             \multirow{10}{*}{200}               & Fixed, $\alpha = 0.001$ & 84.46                 & 15.41                             & 85.28                 & 12.98                           \\
				                                                 & Fixed, $\alpha = 0.01$  & 92.53                 & 7.34                              & 92.65                 & 5.61                            \\
				                                                 & Fixed, $\alpha = 0.1$   & 97.72                 & 2.15                              & 97.08                 & 1.18                            \\
				                                                 & Fixed, $\alpha = 1$     & 10.07                 & 89.8                              & 9.89                  & 88.37                           \\
				                                                 & Fixed, $\alpha = 10$    & 10.02                 & 89.85                             & 9.84                  & 88.42                           \\
				                                                 & Cosine, $\alpha = 0.1$  & 96.42                 & 3.45                              & 96.11                 & 2.15                            \\
				                                                 & Cosine, $\alpha = 1$    & 19.14                 & 80.73                             & 19.07                 & 79.19                           \\
				                                                 & GOLS-I                  & 94.56                 & 5.31                              & 94.6                  & 3.66                            \\
				                                                 & GOALS-4                 & 98.8                  & 1.07                              & 97.53                 & 0.73                            \\
				                                                 & GOS                     & \textbf{99.87}        & 0                                 & \textbf{98.26}        & 0                               \\ \hline
				             \multirow{10}{*}{1000}              & Fixed, $\alpha = 0.001$ & 84.42                 & 15.51                             & 85.29                 & 13.08                           \\
				                                                 & Fixed, $\alpha = 0.01$  & 92.39                 & 7.54                              & 92.65                 & 5.72                            \\
				                                                 & Fixed, $\alpha = 0.1$   & 97.79                 & 2.14                              & 97.12                 & 1.25                            \\
				                                                 & Fixed, $\alpha = 1$     & 10.31                 & 89.62                             & 10.14                 & 88.23                           \\
				                                                 & Fixed, $\alpha = 10$    & 10.06                 & 89.87                             & 9.94                  & 88.43                           \\
				                                                 & Cosine, $\alpha = 0.1$  & 96.77                 & 3.16                              & 96.45                 & 1.92                            \\
				                                                 & Cosine, $\alpha = 1$    & 18.5                  & 81.43                             & 18.52                 & 79.85                           \\
				                                                 & GOLS-I                  & 94.55                 & 5.38                              & 94.57                 & 3.8                             \\
				                                                 & GOALS-4                 & 96.51                 & 3.42                              & 96.18                 & 2.19                            \\
				                                                 & GOS                     & \textbf{99.93}        & 0                                 & \textbf{98.37}        & 0                               \\ \hline
				      \multirow{10}{*}{Overall, $ R_{y} $}       & Fixed, $\alpha = 0.001$ & \multicolumn{1}{c}{-} & 58.08                             & \multicolumn{1}{c}{-} & 49.91                           \\
				                                                 & Fixed, $\alpha = 0.01$  & \multicolumn{1}{c}{-} & 26.66                             & \multicolumn{1}{c}{-} & 20.74                           \\
				                                                 & Fixed, $\alpha = 0.1$   & \multicolumn{1}{c}{-} & 6.27                              & \multicolumn{1}{c}{-} & \textbf{3.47}                   \\
				                                                 & Fixed, $\alpha = 1$     & \multicolumn{1}{c}{-} & 355.57                            & \multicolumn{1}{c}{-} & 350.93                          \\
				                                                 & Fixed, $\alpha = 10$    & \multicolumn{1}{c}{-} & 355.93                            & \multicolumn{1}{c}{-} & 351.67                          \\
				                                                 & Cosine, $\alpha = 0.1$  & \multicolumn{1}{c}{-} & 10.82                             & \multicolumn{1}{c}{-} & 6.91                            \\
				                                                 & Cosine, $\alpha = 1$    & \multicolumn{1}{c}{-} & 285.57                            & \multicolumn{1}{c}{-} & 280.22                          \\
				                                                 & GOLS-I                  & \multicolumn{1}{c}{-} & 29.03                             & \multicolumn{1}{c}{-} & 23.81                           \\
				                                                 & GOALS-4                 & \multicolumn{1}{c}{-} & 12.49                             & \multicolumn{1}{c}{-} & 10.24                           \\
				                                                 & GOS                     & \multicolumn{1}{c}{-} & \textbf{6.05}                     & \multicolumn{1}{c}{-} & 5.43                            \\ \hline
			\end{tabular}}
	\end{center}
	\caption{\revOne{Top average training and testing accuracies over the ten runs for the SGD optimizer with various learning rate strategies, $ y $, including the fixed learning rates, cosine annealing with warm restart, GOLS-I, GOALS-3, and GOS on the N-II architecture with different batch sizes, $ |\mathcal{B}| = $ 10, 100, 200, 1000. It also measures the overall relative robustness, $ R_{y} $, by computing the difference, $ \psi_{y,|\mathcal{B}|} $, between the performance and the best one from the different batch sizes. The highest train, test accuracies, and the lowest robustness measures are indicated in bold.}}
	\label{mnist_table1}
\end{table}

\section{Conclusions}
Dynamic mini-batch sub-sampling causes point-wise discontinuous loss functions, making function value minimization impractical for line search, as local minima may be identified at discontinuities throughout the sampled domain. However, we extend the vanilla gradient-only surrogate (GOS), which employs only directional-derivative information, and empower it by curvature condition to propose our gradient-only approximation line search (GOALS). Learning rates are determined by approximating the location of SNN-GPPs using directional derivative information in quadratic approximation models.

The simplicity of the quadratic model increases computational efficiency but introduces a bias. We also proposed a bracketing strategy using the Regula-Falsi method for constructing approximations. The strategy looks for domains containing SNN-GPPs, using a modified strong Wolfe condition. Furthermore, users can change the degrees of undershooting and overshooting by altering the curvature parameter depending on the line search's preferred behavior. The choice may be dependent on the search direction.

We have tested GOALS against GOS for ResNet-18 and EfficientNet-B0 with the CIFAR-10 dataset to resolve learning rates of the search directions of popular optimizers such as SGD, \textsc{RMSprop}, and \textsc{Adam}. Although the test results showed that GOS without convergence proof tends to outperform GOALS with convergence proof when the search direction is SGD or \textsc{RMSprop} due to restrictions, GOALS outperforms GOS when the chosen search direction is \textsc{Adam}. \revOne{Additionally, we chose the best hyperparameter settings for GOALS based on the robustness criterion we proposed. }

%
%

\begin{acknowledgements}
This research was supported by the National Research Foundation (NRF), South Africa and the Center for Asset Integrity Management (C-AIM), Department of Mechanical and Aeronautical Engineering, University of Pretoria, Pretoria, South Africa. We would like to express our special thanks to Nvidia Corporation for supplying the GPUs on which this research was conducted.
\end{acknowledgements}

%
%

\bibliographystyle{spbasic}      
\bibliography{gosbib}   

\end{document}